\documentclass[letterpaper, 10 pt, conference]{ieeeconf}  % 
\IEEEoverridecommandlockouts       
\usepackage{xspace}
\usepackage[table]{xcolor}
\usepackage{graphicx}
\usepackage{amsmath}
\usepackage{amssymb}  % assumes amsmath package installed
\usepackage{bibunits}
\usepackage{booktabs}
\usepackage{url}
\usepackage{tabularx}
\usepackage[linesnumbered,ruled,vlined]{algorithm2e}
\usepackage{algpseudocode}
\usepackage[cachedir=mintedcache]{minted}

\newcommand{\eg}{\textit{e.g.,}\xspace}

\newcommand{\ie}{\textit{i.e.,}\xspace}
\newcommand{\etc}{\textit{etc.}\xspace}

\usepackage[most]{tcolorbox}
\newcommand{\shortskip}{\vspace{1mm}}
\newcommand{\medkip}{\vspace{3mm}}

\title{\LARGE \bf
Heterogeneous Robot Collaboration in Unstructured Environments \\with Grounded Generative Intelligence
}

\author{
Zachary Ravichandran, Fernando Cladera, Ankit Prabhu, Jason Hughes, \\Carlos Nieto-Granda, Varun Murali, Camillo J. Taylor, George J. Pappas, and Vijay Kumar\thanks{V. Murali is with the Dept. of Electrical and Computer Engineering at Texas A\&M University. 
C. Nieto-Granda is with the U.S. Army Research Laboratory.
All other authors are with the GRASP Laboratory at the University of Pennsylvania. Correspondence can be addressed to \texttt{zacravi@seas.upenn.edu}. 
We acknowledge support from ARL DCIST CRA W911NF-17-2-0181, NSF Grant CCF-2112665, and the NSF Graduate Research Fellowship Program.
}
}

\begin{document}

\maketitle
\thispagestyle{empty}
\pagestyle{empty}

%%%%%%%%%%%%%%%%%%%%%%%%%%%%%%%%%%%%%%%%%%%%%%%%%%%%%%%%%%%%%%%%%%%%%%%%%%%%%%%%
\begin{abstract}
While heterogeneous teams have typically been designed for well-specified missions with known semantics,
generative intelligence, i.e., large language models (LLMs) and vision language models (VLMs), opens the possibility of teams that infer mission-relevant semantics and subtasks given high-level natural language specifications and environmental context.
However, current LLM- and VLM-enabled teaming methods typically assume well-structured and known environments, limiting performance in complex real-world settings.
We address these limitations via \textsc{SPINE-HT}, a framework that grounds the reasoning abilities of LLMs in the evolving context of a heterogeneous robot team through a three-stage process.
Given mission specifications and team capabilities in natural language, an LLM infers necessary subtasks.
These subtasks are validated for feasibility and then assigned to robots based on capabilities such as traversability or perception. 
Finally, the framework uses feedback---semantic maps and task outcomes---acquired by the robots to update subtasks online.
In simulation experiments with closed-loop perception and control,
our framework achieves nearly twice the success rate  compared to prior LLM-enabled heterogeneous teaming approaches.
In real-world experiments with 
a Clearpath Jackal, a Clearpath Husky, a Boston Dynamics Spot, and a high-altitude UAV, our method achieves an 87.5\% success rate
in missions requiring reasoning about robot capabilities and refining subtasks with online feedback.
More information is provided at: \url{https://zacravichandran.github.io/SPINE-HT}.
\end{abstract}

\section{Introduction}
\label{sec:intro}

\begin{figure}[t!]
    \centering
    \includegraphics[width=1.0\linewidth]{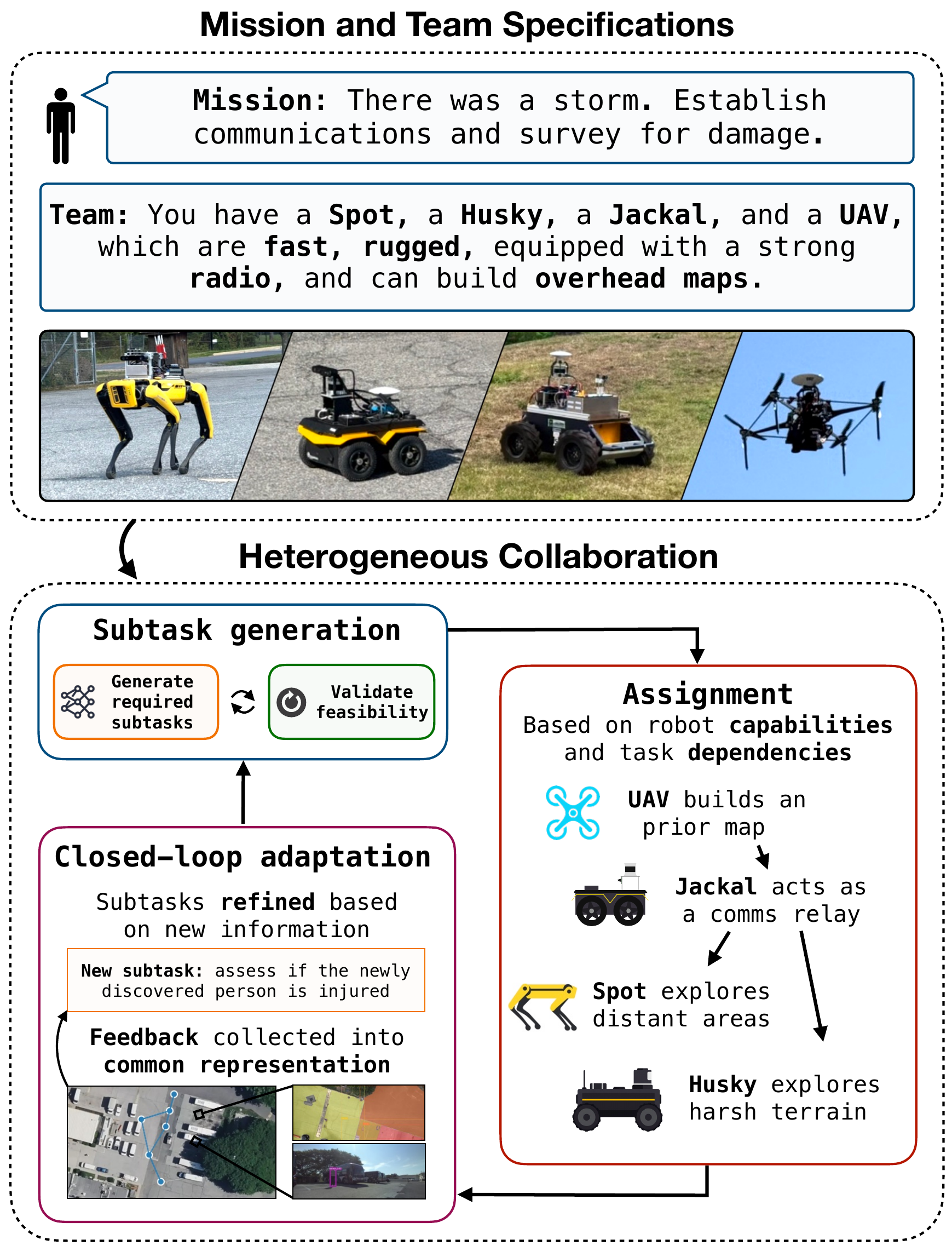}
    \caption{We present a method for heterogeneous robot collaboration that infers semantics and subtasks based on natural language mission specifications, assigns them to robots based on capabilities and subtask dependencies, then updates them as information is acquired online.
    \vspace{-12pt}
    }
    \label{fig:intro}
\end{figure}

Consider a heterogeneous robot team operating after a heavy storm. 
A user may provide the mission: \emph{There was a storm. Establish communications and survey for damage}, which is an incomplete specification as it omits relevant semantics and subtasks.
Effective collaboration requires a planner that first infers mission-relevant semantics (e.g., people, damaged infrastructure) and infers dependency-aware subtasks (e.g., looking for people requires mapping the environment), while leveraging unique robot capabilities (e.g., rugged robots traverse debris, UAVs build overhead maps).
Because missions occur in unstructured environments where prior maps are unavailable and robots operate amidst hazards,
the planner must adapt subtasks based on feedback acquired online, such as map updates or subtask results.
While recent methods have advanced this vision, significant gaps remain.

Multi-robot teaming has traditionally been addressed via optimization-based methods, where goals are encoded in objective functions and team capabilities are described via constraints~\cite{prorok2017impact, prorokspeciesmultirobo}.
Optimization-based approaches have addressed large-scale collaboration problems such as distributed mapping, and they offer a principled approach to task allocation.
However, these methods typically assume well defined semantics, a well-specified objective function, and \textit{a priori} knowledge of the environment, limiting their applicability in large-scale and unstructured settings.

\begin{figure*}[t!]
\vspace{3pt}
    \centering
    \includegraphics[width=0.95\linewidth]{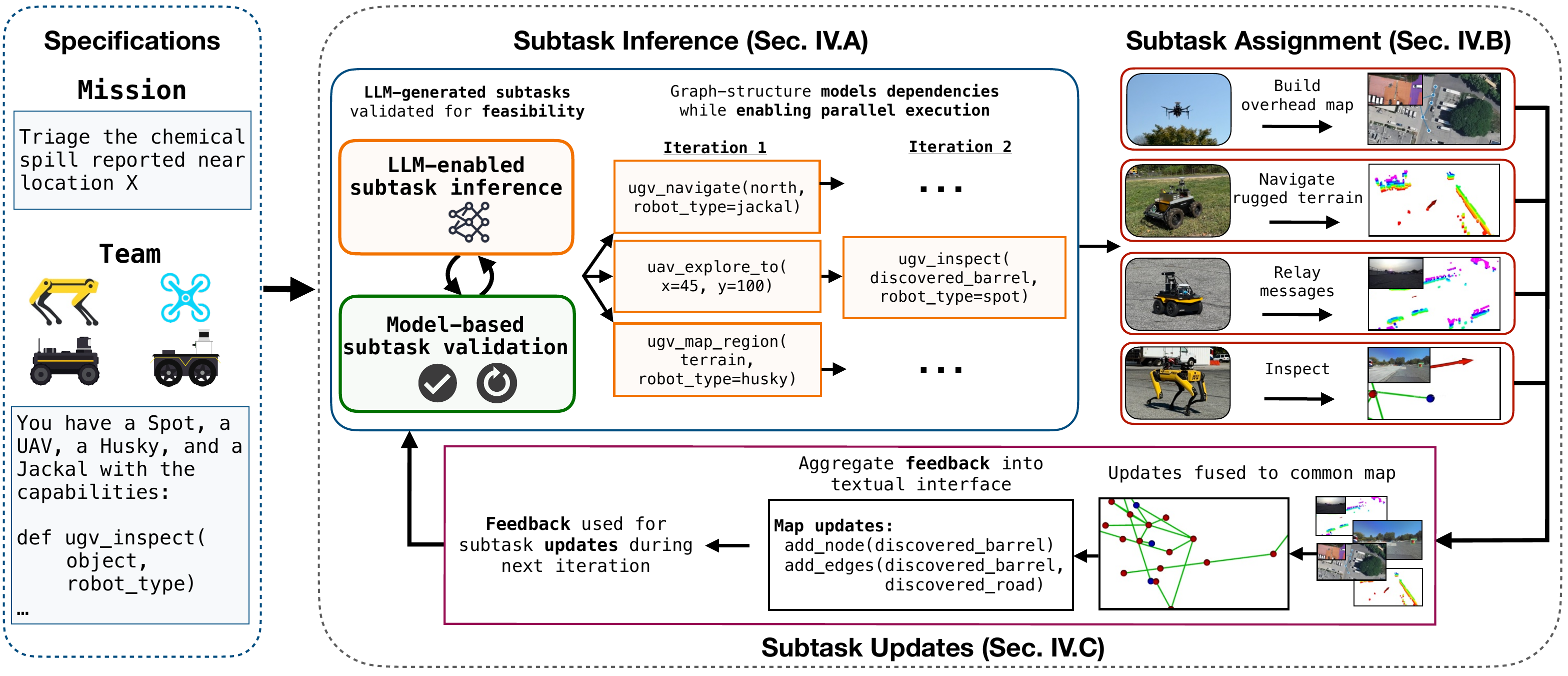}
    \caption{Our planner takes as input natural language mission specifications  and team capabilities. The planner then uses an LLM to infer grounded subtasks with their dependency ordering (\S\ref{sec:method_decomp}). 
    Subtasks are then assigned to robots based on capabilities (\S\ref{sec:method_assign}). Feedback comprising semantic map updates and subtask outcomes are collected and used for subtask updates during subsequent iterations (\S\ref{sec:method_feedback}).\vspace{-6pt}}
    \label{fig:method}
\end{figure*}

A second line of multi-robot teaming research focuses on deployment in unstructured environments~\cite{agha2022nebula, miller2020mine, miller2024spomp, chang2022lamp}.
The success of these systems largely comes from an emphasis on closed-loop operation---the planner is not considered in isolation, rather it is deployed in-the-loop with perception and inter-robot communication.
While these approaches improve robustness, they continue to rely on predefined semantic categories and well-defined missions with explicit task decompositions.
For example, a planner may be designed to find all vehicles in a given area, but adapting it to search for paths in a cave requires core algorithmic changes~\cite{miller2020mine, miller2024spomp}.

The increasing maturity of generative intelligence promises to endow multi-robot teams with the ability to infer relevant semantics and subtasks given language-specified missions.
However, LLMs are prone to hallucination and lack grounding in physical constraints, making reliable deployment in unstructured settings an open challenge.
While several LLM-enabled approaches have demonstrated multi-robot teaming given language specifications, existing work has primarily been designed for structured environments such as simulators or laboratories, relying on known maps and lacking robust mechanisms for validating that generated plans are physically realizable or incorporating perceptual feedback~\cite{kannan2023smart, chen2024scalablemultirobotcollaborationlarge, liu2025coherentcollaborationheterogeneousmultirobot, zhu2025dexterllmdynamicexplainablecoordination}. 

To bridge this gap, we propose \textsc{SPINE-HT}, a framework that grounds LLM-enabled reasoning in the physical and semantic context of heterogeneous robot teams. 
As illustrated in Fig.~\ref{fig:method}, given mission and team specifications in natural language, an LLM infers subtasks with mission-relevant semantics that are validated for feasibility, mitigating the risk of hallucinated or infeasible plans. Subtasks are structured as a directed acyclic graph (DAG) to model dependencies, assigned to robots via constrained optimization over capabilities, and updated online using feedback from semantic mapping and task execution.
To summarize, we propose a planner that:
\begin{enumerate}
    \item[1.] Infers a dependency-aware graph of grounded subtasks from incomplete natural language specifications, capturing mission-relevant semantics while respecting robot capabilities and environmental constraints.
    
    \item[2.] Allocates subtasks  while respecting subtask dependencies and heterogeneous platform capabilities.

    \item[3.] Enables closed-loop subtask updates by using feedback from semantic mapping and execution outcomes.
\end{enumerate}
We evaluate these contributions via simulated and real-world experiments with closed-loop mapping and control. 
In real-world experiments, our framework achieves a mission success rate of 87.5\%, while producing plans that are within 15\% of expert efficiency in terms of subtasks required.
Compared to state-of-the-art LLM-enabled heterogeneous planning approaches, our method achieves nearly twice the mission success rate and subtask efficiency.

\section{Related work}

Heterogeneous robot teaming has been addressed via optimization and formal methods, robust systems for deployment in unstructured environments, and most recently generative intelligence via LLM-enabled methods.

\medskip

\noindent \textbf{Optimization and formal methods.}
Heterogeneous robot planning has traditionally been formulated as an optimization problem, where tasks are encoded via objective functions and constraints govern which robot may be assigned to a given task~\cite{prorok2017impact}. 
For example, a coverage control problem may yield an objective that specifies robot locations, and constraints may encode robot capabilities such as its sensing modalities or speed, and this formulation admits tractable solutions via constrained optimization approaches~\cite{prorokspeciesmultirobo}.
By incorporating further information such as robot dynamics, optimization approaches may be employed to jointly solve task allocation and motion planning~\cite{messing2022grstaps}.
\cite{gosrich2023multi} employs graph-based representations to model inter-robot task dependencies, which enables multi-robot allocation over complex tasks. 
Formal methods specifications, such as linear temporal logic (LTL) or Planning Domain Definition Language (PDDL), have also been used to represent multi-robot missions~\cite{Kalluraya2023} and provide guarantees on correctness and feasibility. 
These approaches require an expert user to explicitly encode robot capabilities, detailed mission specifications including task dependencies, and assume a known world model. 
\medskip

\noindent\textbf{Multi-robot systems in unstructured environments}. 
While the above methods typically consider well-structured environments, a line of work develops multi-robot teaming approaches for unstructured environments, where the world is unknown, the terrain is hazardous, and the robots may not enjoy reliable communication~\cite{cladera2024challenges, ravichandran2025deploying, fadhil2024capability}.
These systems have typically been designed for missions requiring exploration and object search, with teams comprising air and ground robots~\cite{miller2020mine, miller2024spomp, agha2022nebula}. 
The success of these systems largely relies on planners embedded in robust feedback loops with perception, control, inter-robot communication, and other key modules~\cite{tranzatto2022cerberus, cladera2024enabling, chang2022lamp}.
While the ability of these systems to operate in large-scale and unstructured environments is impressive, these systems require missions with predefined semantic categories and explicit task decompositions, and adapting to new specifications requires development by an expert user. 

\medskip

\noindent\textbf{Generative intelligence for robot teams.}
Recent work leverages the generative intelligence of LLMs to translate linguistic mission and robot specifications for multi-robot collaboration. 
Some approaches focus on homogeneous robot teams in centralized or decentralized manners~\cite{chen2024scalablemultirobotcollaborationlarge, roco}.
SMART-LLM~\cite{kannan2023smart} and similar works consider heterogeneous robot collaboration~\cite{zhu2025dexterllmdynamicexplainablecoordination, gupta2025generalizedheterllm}. 
The proposed methods use an LLM to decompose language specifications into robot-specific tasks, but they require fully-known world models.
COHERENT furthers LLM-enabled heterogeneous robot collaboration by incorporating feedback mechanisms for plan validation and updated world models~\cite{liu2025coherentcollaborationheterogeneousmultirobot}. 
However, this method does not model task dependencies, assumes simplistic world models, and its feedback mechanisms rely on auxiliary LLMs that are also prone to hallucination, without model-based validation to check feasibility.
\cite{strader2025languagegroundedhierarchicalplanningexecution} uses an LLM to translate language commands into a PDDL plan in a grounded world model.
However, their method requires a fully-known map and is limited to coordination with fully-specified goals (\ie a user gives specific instructions to each robot).
\cite{tfr} recently proposed a method for air-ground robot teaming with language-specified missions, but their method is not scalable to different robot teams.
In summary, existing LLM-enabled heterogeneous robot teaming methods lack robust mechanisms for \emph{grounding} semantic goals in robot capabilities and physical constraints, preventing operation in unstructured environments.

\section{Problem Definition}
\label{sec:problem}

We consider a team of $R$ heterogeneous robots, where each robot is equipped with a set of high-level capabilities such as mapping, inspection, or exploration.
The team is provided with an incomplete mission specification in natural language (e.g., ``survey for damage'') that does not specify which semantic entities are relevant, what subtasks are required, or how they should be allocated across the team. 
During operation, each robot constructs a local semantic map, and these are aggregated into a shared map $\mathcal{M}$ comprising task-relevant metric-semantic information.
We assume that the robots are equipped with open-set semantic mappers that may be configured by the planner.

Our goal is to design a planner that iteratively: (1) infers mission-relevant semantics and generates appropriate subtasks $\mathcal{T}$; these subtasks invoke capabilities given the current map (e.g., where to explore or what to inspect); (2) assigns subtasks to robots based on their capabilities and constraints; and (3) updates subtasks as $\mathcal{M}$ and task outcomes evolve.

\section{Heterogeneous Collaboration}

\textsc{SPINE-HT} addresses the problem defined in \S\ref{sec:problem} through a three-stage framework that operates in closed-loop fashion, as illustrated in Fig.~\ref{fig:method}.
First, the \emph{subtask inference} module maps language specifications into a DAG of grounded subtasks,
explicitly modeling inter-task dependencies while enabling parallel execution where possible (\S\ref{sec:method_decomp}).
Next, the \emph{subtask assignment} module allocates tasks to robots based on their capability, considering both hard constraints (\eg physical feasibility) and semantic capabilities stated in the language specification (\S\ref{sec:method_assign}).
Finally, the subtask update module incorporates execution feedback---semantic map updates and task outcomes---to adapt subtask goals as the team's knowledge of the environment evolves (\S\ref{sec:method_feedback}).

\subsection{Subtask Inference}
\label{sec:method_decomp}

The subtask inference module maps natural language specifications into grounded subtasks via LLM-guided reasoning and formal verification.
This process comprises three key stages---providing input specifications, generating subtasks, and subtask validation.
\medskip

\noindent \textbf{Input specifications.} The subtask inference module is provided with incomplete mission and team specifications which describe robot capabilities. 
Each capability takes two types of arguments: grounding arguments describing metric and semantic goals and platform arguments describing the type of robot required. 
For example, the \verb|ugv_inspect| capability:
\begin{tcolorbox}[colback=gray!3, colframe=black,left=1mm, right=1.5mm, top=1.5mm, bottom=1mm] \footnotesize
\begin{minted}[breaklines, breakanywhere]{python}
def ugv_inspect(object: str, query: str, robot_type: Literal["jackal", "spot"]):
"""Inspect `object` for information described in `query.` Robots "jackal" and "spot" are equipped with RGB-D cameras. """
\end{minted}
\end{tcolorbox}
\noindent defines the grounding arguments \verb|object| (target entity) and \verb|query| (semantic information to gather). 
The platform argument \verb|robot_type| constrains execution of that capability to  a robot equipped with an RGB-D camera. 
Platform arguments may be either a robot type (e.g., ugv) or specific robot, which enables soft capability specifications (e.g., ``\verb|spot| has the best camera in the team'').

\medskip
\noindent \textbf{Generating subtasks.} Given the input specifications, the subtask inference LLM produces a DAG, where nodes consist of grounded subtasks and edges denote inter-task dependencies (see Fig.~\ref{fig:method}).
The DAG structure identifies independent subtasks while respecting dependency constraints---tasks connected by a directed path must run sequentially, while independent components may run concurrently. 
To improve plan quality and interpretability, the LLM employs chain-of-thought reasoning (CoT), 
explicitly justifying key steps in its subtasks generation process~\cite{cot_llm}.

\medskip

\noindent \textbf{Subtask validation.} 
While LLMs enable powerful contextual reasoning, their potential to hallucinate information may lead to infeasible plans.
Recent work has demonstrated that embedding LLM-enabled planning with model-based validation ensures generated plans are physically realizable~\cite{ravichandran_spine,tfr}.
We therefore adopt the approach of \cite{ravichandran_spine} and utilize an Assume-Guarantee validation framework, where each capability defines preconditions that must hold for invocation~\cite{Giannakopoulou2004AssumeGuarentee}.
For example, the \verb|ugv_inspect| capability discussed above requires that the target \verb|object| is visible and reachable in the current map. 
If this capability is invoked with an unreachable object, the LLM would receive the feedback ``\verb|object| is unreachable, and you may need to find a path.''
This iterative generation-validation feedback loop runs until all subtasks satisfy their defined preconditions, ensuring feasibility before subtasks are assigned to robots.
Tab.~\ref{tab:behaviors} summarizes key capabilities available to the planner along with their primary preconditions.

\begin{table}[]
  \vspace{8pt}
    \centering
    \scriptsize
    \begin{tabular}{c|p{2.3cm} p{2cm} }
        \toprule
        Capability & Description & Constraints \\
        \midrule
        \verb|ugv_map_region| & augment semantic map & region within map \\
        \verb|ugv_inspect| & inspect object & object visible \\
        \verb|ugv_navigate| & go to a known location & path exists\\        
        \verb|uav_explore_to| & discover new region & metric goal\\
        \bottomrule
    \end{tabular}
    \caption{Subset of capabilities considered in our work.
    \vspace{-24pt}
    }
    \label{tab:behaviors}
\end{table}

\subsection{Subtask Assignment}
\label{sec:method_assign}
The subtask assignment module assigns validated subtasks from the mission graph to specific robots through constrained optimization, accounting for heterogeneous capabilities and minimizing execution costs.
At each planning iteration, given the set of generated subtasks $\mathcal{T}$, the module constructs a candidate subtask set $\mathcal{T'} \subseteq \mathcal{T}$ 
comprising tasks ready for assignment, \ie dependency-free nodes.

The assignment problem is formulated as a linear optimization over the binary decision variable $X \in \{0, 1\}^{R  \times \lvert \mathcal{T'} \rvert }$, where $R$ is the number of robots and $X_{rt} = 1$ indicates that robot $r$ is assigned task $t$.
The cost matrix $C \in \mathbb{R}^{R \times \lvert \mathcal{T'} \rvert}$ encodes capability constraints and additional tasking preferences. $C_{rt}$ 
is infinite if robot $r$ is not capable of completing task $t$\footnote{In practice, we set infinite costs to 1k.}; otherwise, the cost is the traversal distance required for the robot to complete the task. 
This yields:
\begin{align}
    \min_X \quad & \sum_{r=1}^R\sum_{t=1}^{\lvert \mathcal{T'} \rvert} C_{rt} X_{rt}  \\
    \text{s.t.}\quad & \sum_{t=1}^{\lvert \mathcal{T'} \rvert} X_{rt} = 1 \quad \forall r \in \{1, \dots, R\} \\
        & \sum_{r=1}^{R} X_{rt} \leq 1 \quad \forall t \in \{1, \dots, \lvert \mathcal{T'} \rvert \}
\end{align}
The first constraint ensures that all robots are assigned a task, while the second ensures that tasks are assigned to at most one robot.
To handle cases where $\lvert \mathcal{T}' \rvert < R$, we introduce ``idle'' tasks with costs such that, if a robot cannot achieve any candidate task, it will be assigned as ``idle.'' Conversely, unassigned tasks are kept as candidates for the next assignment iteration.
The assignment process operates iteratively as the mission progresses. Upon completion, subtasks are marked as complete in the DAG, thereby creating an updated active subgraph where previously blocked tasks may be available for assignment.
A new candidate subtask set $\mathcal{T'}$ is computed and assigned to the robots, and this process iterates until all tasks in the mission graph have been assigned.

\subsection{Subtask Updates}
\label{sec:method_feedback}

Robots continuously gather information about the world during mission execution.
The subtask update module collects this information, synthesizes it into a structured update, and provides it to the subtask inference module (\S\ref{sec:method_decomp}) for the next iteration of planning.
This closed-loop operation is essential for robust operation in unstructured environments, where environmental knowledge is incomplete and constantly evolving.
We consider two primary sources of feedback: semantic map updates and subtask outcomes. 
We first describe the semantic map representation used by our planner and then describe feedback representations.

\medskip

\noindent\textbf{Online open-set semantic mapping.}
\textsc{SPINE-HT} models the world via a topological graph-based semantic map---a rich representation that has facilitated a variety of semantic planning methodologies~\cite{ravichandran_spine, tfr, strader2025languagegroundedhierarchicalplanningexecution}.
The map consists of two node types: \textit{regions} and \textit{objects}.
Region nodes indicate traversable locations in freespace, while object nodes correspond to semantic entities in the environment. 
Map connectivity is defined by two edge types: region-to-region edges define traversable robot paths, while object-to-region edges denote a visibility relationship between an object and point in freespace. Importantly, our mapping framework enables \emph{open-set semantics} via open-set object detection and vision-language models (VLMs), which may enrich nodes with open-ended descriptions (\eg ``this car has a dent on its side'').
In contrast to planning methods that assume fixed semantics at planning time~\cite{strader2025languagegroundedhierarchicalplanningexecution,agha2022nebula,prorok2017impact}, our framework must infer and generate mission-relevant semantics.

\begin{figure}
\vspace{5pt}
    \centering
    \includegraphics[width=1.0\linewidth]{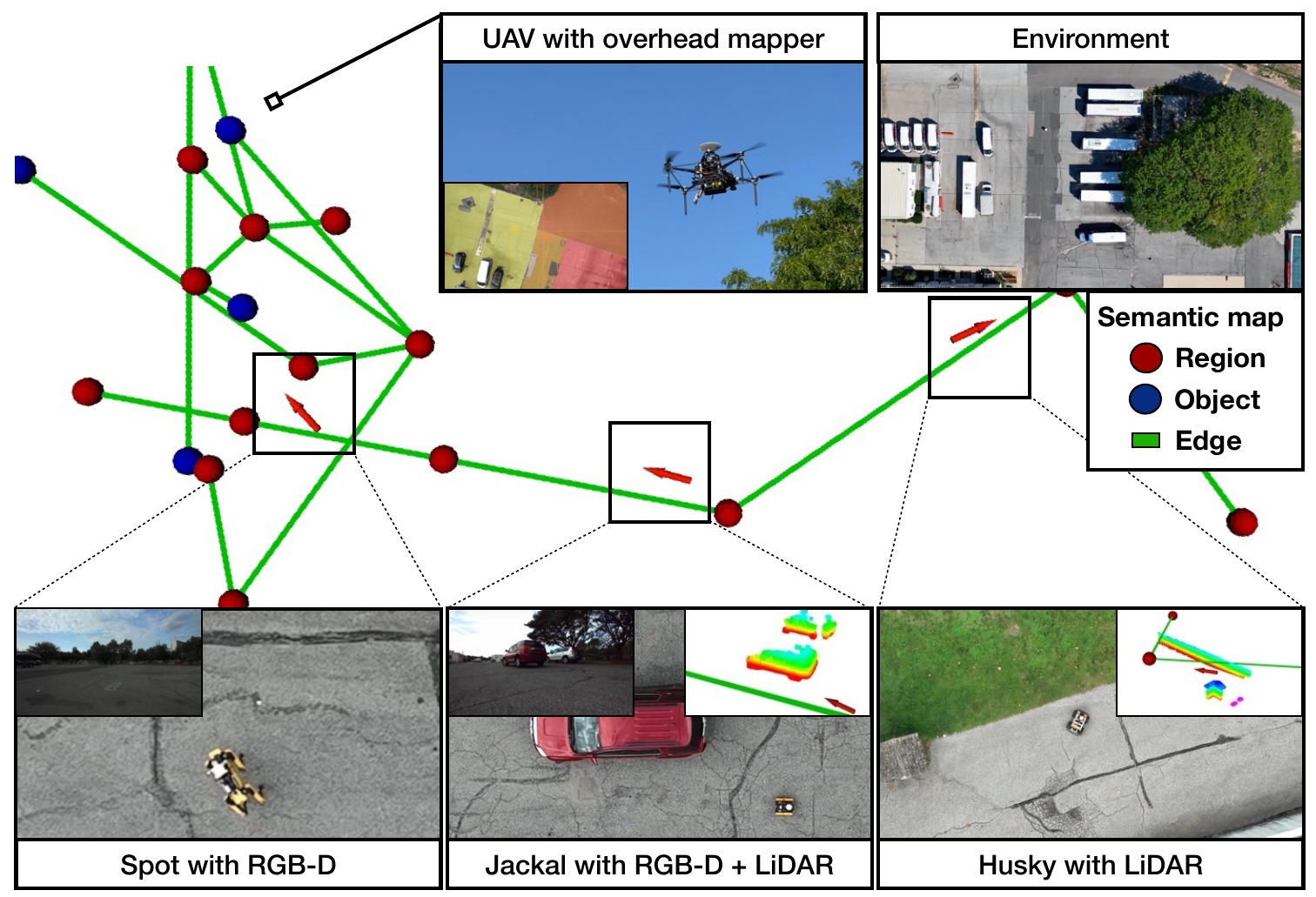}
    \caption{Our framework aggregates updates from  heterogeneous robots into a common semantic map used for subtask refinement. %\fc{See comment on shared map.}
    \vspace{-6pt}
    }
    \label{fig:mapping}
\end{figure}

While \textsc{SPINE-HT} accepts a prior semantic map (\eg from aerial imagery or a UAV), each robot constructs a local map online with sensory data (\eg RGB-D images, LiDAR). 
The sparse, graph-based nature of these semantic maps facilitates online aggregation across varying sensor modalities during operation, where map updates comprise atomic node and edge additions and deletions\footnote{While we assume consistent localization across robots, future work may incorporate advanced distributed mapping techniques that account for mapping conflicts~\cite{miller2024spomp,tian2022kimeramulti}} (see Fig.~\ref{fig:mapping}).

\medskip
\noindent \textbf{Mission feedback and history.}
The feedback module translates map updates and task outcomes into a textual interface that may be ingested by the subtask inference LLM.
Map updates are provided to the LLM via a functional interface that defines core graph operations, such as node and edge additions and removals, as well as more nuanced semantic description. For example, a new \verb|boat| node at location $[10, 20]$ connected to a \verb|dock| would yield the  update:  
\begin{tcolorbox}[colback=white!1, colframe=black,left=1mm, right=1.5mm, top=1.5mm, bottom=1mm] \footnotesize
\begin{minted}[breaklines, breakanywhere]{python}
add_node(name='boat', location=[10, 20], 
description='in good condition',
edges=['dock'])
\end{minted}
\end{tcolorbox}

\noindent Task outcomes are provided via a similar interface that provides key information in a standardized format.
For example, a failed navigation subtask may provide the update ``Subtask \verb|X| failed. path between \verb|node_1| and \verb|node_2| was blocked.'' All updates at a given planning iteration are provided to the subtask inference module via in-context updates.
While storing history via in-context updates leads to accumulating token length, we find that the maximum tokens used in our experiments is roughly 60k, while current LLMs have a context length of over 1M (see \S\ref{sec:sim_result}). 

\section{Experiments}

We conduct experiments to assess \textsc{SPINE-HT}'s ability to fulfill the contributions stated in \S\ref{sec:intro}. Namely, the ability to:
\begin{enumerate}
    \item [\textbf{C1.}] Infer grounded subtasks that capture dependencies.
    \item [\textbf{C2.}] Assign subtasks based on robot capabilities. 
    \item [\textbf{C3.}] Adapt subtasks given feedback acquired online.
\end{enumerate}

\noindent To evaluate our first contribution, we evaluate against missions with \emph{incomplete specifications} that require  the planner to infer relevant subtasks and their dependencies.
To evaluate our second contribution, we design missions requiring heterogeneous collaboration where mission success requires leveraging unique robot capabilities.
Finally, to assess our third contribution, we evaluate in environments where the planner's semantic map is incomplete, requiring the planner to actively explore mission-relevant regions and adapt its plan based on findings acquired online.
We conduct extensive evaluation across both simulation and real-world experiments comprising 40 missions with up to four heterogeneous robot platforms. 
Unlike prior LLM-enabled heterogeneous planning work~\cite{liu2025coherentcollaborationheterogeneousmultirobot, kannan2023smart, zhu2025dexterllmdynamicexplainablecoordination}, we conduct all experiments with closed-loop mapping and control.
This integration stresses the planner's ability to interpret and react to realistic sensor inputs, environmental uncertainties, and execution failures --- all of which are critical for robust deployment in unstructured environments. 
\vspace{-8pt}

\medskip

\begin{table*}[t!]
\vspace{5pt}
    \centering
    \begin{tabular}{c | cccc | cccc |  cccc}
         \toprule
       Method & \multicolumn{4}{c}{Subtask Generation} & \multicolumn{4}{c}{Capability Reasoning} & \multicolumn{4}{c}{Exploration} \\ 
       \cmidrule(lr){2-5}  \cmidrule(lr){6-9}  \cmidrule(lr){10-13} 
       & Suc. ($\uparrow$) & ST ($\downarrow$) & Opt. ($\uparrow$) & Tok. ($\downarrow$) & Suc. ($\uparrow$) & ST($\downarrow$) & Opt.($\uparrow$) & Tok. ($\downarrow$)  & Suc. ($\uparrow$) & ST ($\downarrow$) & Opt. ($\uparrow$) & Tok. ($\downarrow$) \\
       \midrule
       Expert  & 100.0\% & 1.0 & 100.0\% & - & 100.0\% & 1.0 & 100.0\% & - & 100.0\% & 1.0 & 100\% & - \\ 
       LLM baseline     & 37.5\% & 6.2 & 81.0\% & 11.7k & 37.5\% & 5.5 & 87.0\% & 10.2k & 50.0\% & 2.5 & 66.0\% & 8.7k \\
       Ours    & 100.0\% & 1.0 & 90.0\% & 3.5k & 75.0\% & 1.3 & 84.0\% & 5.1k & 100.0\% & 1.2 & 86.0\% & 5.9k\\
        \bottomrule
    \end{tabular}
    \caption{Simulation results across 24 missions.\vspace{-16pt}}
    \label{tab:sim-results}
\end{table*}

\subsection{Experimental Setup}
\label{sec:sec:experiment_setup}
We now provide an overview of the key experimental components: robotic platforms, dimensions of heterogeneity, evaluation environments, baseline methods, mission specifications, performance metrics, and implementation details.

\begin{figure}[t!]
\vspace{5pt}
    \centering
    \includegraphics[width=1.0\linewidth]{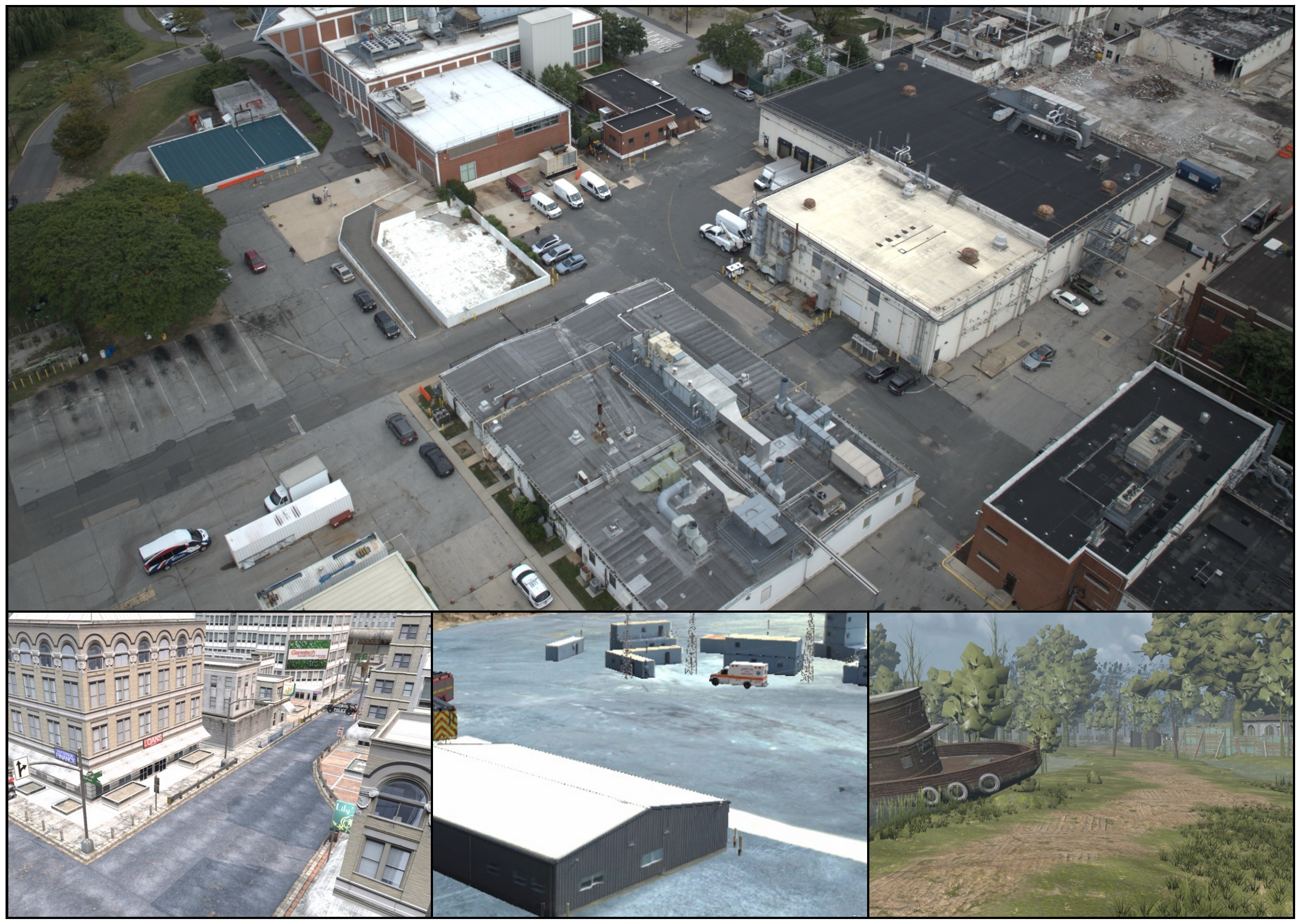}
    \caption{Evaluation environment: real semi-urban office park (top) and three urban, semi-urban, and rural simulation environments (bottom).
    \vspace{-6pt}
    }
    \label{fig:envs}
\end{figure}

\shortskip

\noindent \textbf{Experimental platforms.} We evaluate with a total of four platforms (see Fig.~\ref{fig:intro}).
\emph{Clearpath Jackal}: A small mobile robot  equipped with a LiDAR and RGB-D camera. \emph{Clearpath Husky}: A large, rugged mobile platform capable of traversing obstacles including steep slopes and curbs. It is equipped with LiDAR only, limiting semantic perception. \emph{Boston Dynamics Spot}: Quadrupedal robot with superior mobility compared to wheeled platforms. The Spot is equipped with an RGB-D camera for semantic perception but lacks LiDAR for detailed metric mapping. \emph{High-altitude UAV}: A quadrotor capable of quickly flying at high altitudes (30 meters) but is limited to overhead sensing.

\shortskip

\noindent \textbf{Dimensions of heterogeneity.} Our platforms differ across four key capability dimensions. 
\emph{Perception}: Sensor modalities dictate platform perception capabilities. In our experiments, RGB-D cameras were used for semantic mapping while LiDAR platforms performed metric mapping. 
\emph{Traversability}: Platforms vary in the difficulty of terrain they can traverse and speed at which they move.
\emph{Communication}: Certain robots carry more powerful radios and may support communication within the team.
\emph{Payload}: Certain robots may be equipped with payloads that must be delivered. We vary these dimensions to produce \emph{team configurations}, where robots are given individual capabilities or constraints, \eg a robot may have powerful perception, a robot may be carrying a package, \etc

\shortskip

\noindent \textbf{Baselines.} We compare against the following baselines.
First, we consider an Expert Planner, where a user provides explicit mission specifications comprising step-by-step instructions and optimal robot assignments.
This baseline provides an upper bound estimate on mission performance and emulates existing approaches that require explicit mission instructions and known maps. 
We then consider COHERENT, a state-of-the-art LLM-enabled planner for heterogeneous teaming~\cite{liu2025coherentcollaborationheterogeneousmultirobot}.
COHERENT represents the current LLM-based heterogeneous robot planning paradigm, sharing core characteristics with related methods~\cite{kannan2023smart, zhu2025dexterllmdynamicexplainablecoordination, gupta2025generalizedheterllm} including (1) text-based plan representation, (2) no formal verification, (3) no structured feedback mechanisms to track updates and history throughout a mission. 
To ensure fair comparison, we provide identical capabilities across all methods.

\shortskip

\noindent \textbf{Environments.} We evaluate in both simulation and real-world environments as illustrated in Fig.~\ref{fig:envs}. We use a photorealistic Unity-based simulator featuring large-scale outdoor environments with semantics including factories, offices, communication towers, bridges, and vehicles.
We also conduct experiments in an uncontrolled real-world urban office park with semantics such as vehicles, people, vegetation, and infrastructure.
Crucially, these environments remained active during experiments, forcing the planner to handle unexpected perception inputs and avoid environment hazards.

\shortskip

\noindent \textbf{Mission Specifications.} 
We design missions to emulate real-world scenarios including infrastructure inspection, emergency response, personal assistance, and environmental monitoring.
Mission specifications are provided in natural language and require the planner to both infer relevant subtasks and assign these subtasks based on robot capabilities.
Mission success requires the planner to configure robots in a certain location (e.g., for package delivery or communication) and acquire mission-relevant information (e.g., for infrastructure inspection).
For example, the specification listed in Fig.~\ref{fig:method} requires that the planner look for chemical barrels, people, and other triaging equipment. 
A UAV must map distant areas, while ground robots map closer terrain.
As information is acquired during a mission (e.g., a barrel is found), a ground robot must then perform close-up inspection, and the planner must report findings to the user.

\shortskip

\noindent \textbf{Metrics.} We report four metrics. 
\emph{Success rate:} the percentage of missions achieving all required objectives (Suc.).
\emph{Subtasks:} number of subtasks required for mission completion relative to the expert baseline, for successful missions (ST).
\emph{Optimal subtask assignment:} percentage of subtasks correctly assigned to each robot, as determined by the unique robot capabilities and mission requirements (Opt.).
ST and Opt. are reported relative to the expert planner.
Finally, we report \emph{tokens} required per subtask generation (Tok.).

\shortskip
\noindent \textbf{Implementation details.} We use GPT-4.1 for the subtask generation LLM, and we solve the optimization problem described in \S\ref{sec:method_assign} via linear programming.
We build our robot autonomy off of the implementation proposed in \cite{ravichandran_spine}; we only provide key details, due to space constraints.
The semantic mapping framework takes three sources of input: odometry, a pointcloud, and RGB-D imagery. 
We use Direct Lidar Inertial Odometry for localization~\cite{chen2022dlio}.
Pointclouds are segmented for obstacles by GroundGrid~\cite{GroundGrid}, and these obstacles are used to determine freespace for region and edge placement.
GroundingDino  performs open-vocabulary object detection~\cite{liu2023groundingdino}.
Detections are localized in 3D space, associated via a multiple-hypothesis tracker~\cite{reid1979algorithm}.
Finally, we use the VIP-Llava VLM to describe regions and objects~\cite{cai2023makingllava}. 
We implement high-altitude UAV mapping based on the system proposed in \cite{tfr}.
Our framework assumes intermittent communication~\cite{cladera2024enabling},
which requires only a basestation responsible for running subtask generation and allocation (\S\ref{sec:method_decomp}, \S\ref{sec:method_assign}) to maintain internet connectivity, while robots may drop in and out of communication range during the mission.
We implement robot navigation with ROS2 Nav2, which handles point navigation and collision avoidance\footnote{https://docs.nav2.org/}.

\subsection{Simulation Results}
\label{sec:sim_result}

We run a total of 24 missions requiring between 3 and 7 subtasks.
We report simulation results  across the three mission categories---subtask generation, capability reasoning, and exploration---in Tab.~\ref{tab:sim-results}.
Our method achieves an average success rate of 100.0\%, 75.0\%, and 100.0\% across these three mission categories respectively, while the LLM baseline achieves 37.5\%, 37.5\%, and 50.0\%.
The LLM baseline also required more subtasks to complete missions (between 2.1 and 6.2 times) and used more tokens (up to 3.3 times).
While the subtask optimality of the LLM baseline was comparable to our method for the first two mission categories, it was  nearly 20\% lower for the exploration missions.

We find that while the LLM baseline produced reasonable semantic goals, it suffered from two primary failure modes. 
First, the baseline struggled to efficiently ground plans in the environment, often composing repetitive and suboptimal navigation-related subtasks. 
These unnecessarily verbose subtasks would increase the planning horizon, and often lead to hallucination.
Interestingly, this problem was alleviated during exploration missions where the prior map received by the planner was significantly smaller and the LLM  baseline was able to leverage the available capabilities to explore mission-relevant locations.
In contrast, \textsc{SPINE-HT} generally produced targeted subtasks that identified key locations in the environment across all three mission scenarios.
The LLM baseline also struggled to correctly invoke perception capabilities.
For example, the LLM baseline often did not properly interpret the results of a VLM query, or it would attempt to invoke  queries on robots without VLMs.

\begin{figure}[t!]
\vspace{5pt}
    \centering
    \includegraphics[width=1.0\linewidth]{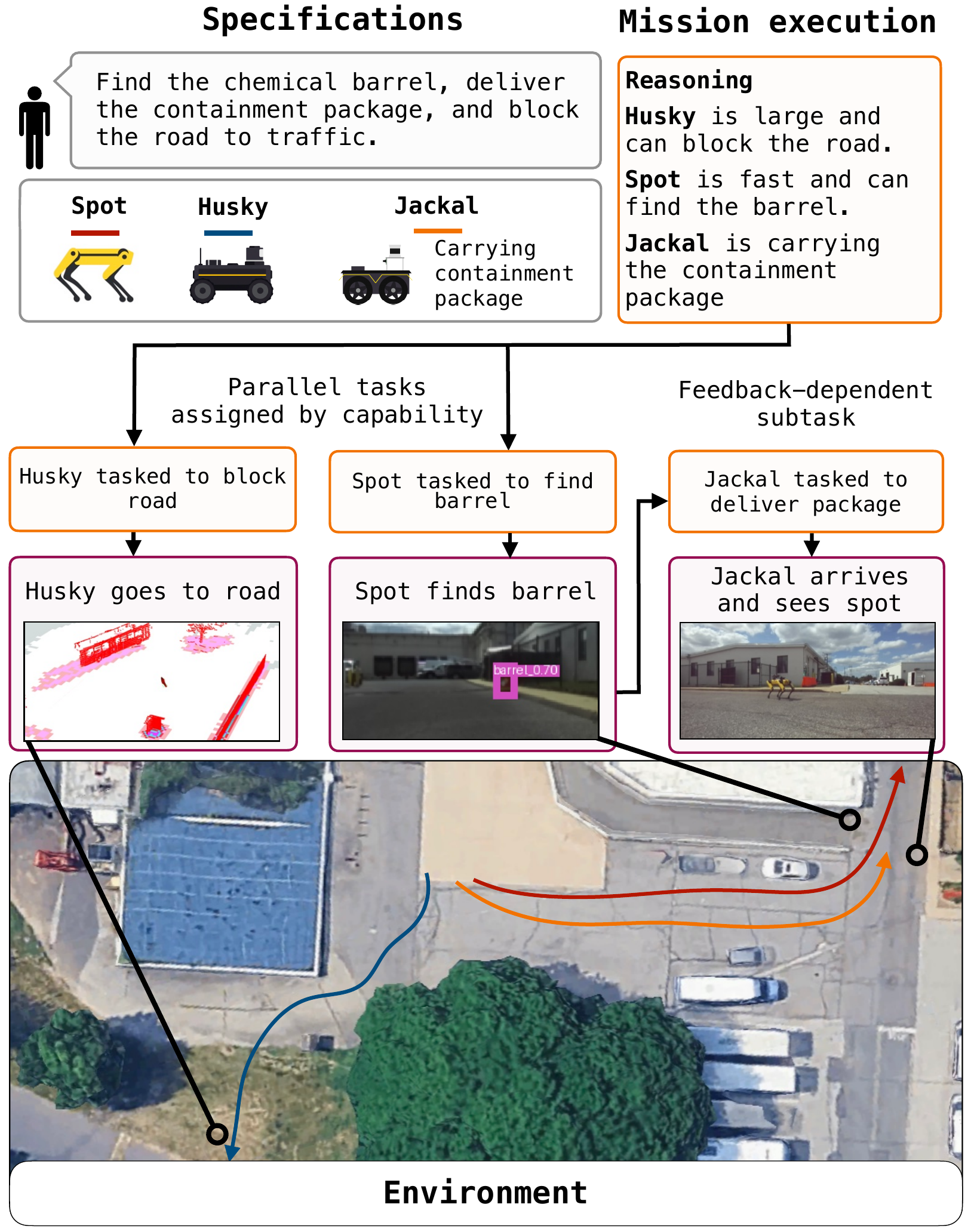}
    \caption{Real-world result where the planner must infer grounded subtasks, reason about robot capabilities, and refine subtasks online given feedback.
    \vspace{-16pt}
    }
    \label{fig:example_real_mission}
\end{figure}

\subsection{Real-world Results}
\label{sec:real_results}
We ran a total of 16 real-world experiments requiring at least 4 and up to 9 subtasks with teams between 2 and 4 robots, and we report results in Tab.~\ref{tab:real-results}.
\textsc{SPINE-HT} achieved a mission success rate of 87.5\%, while requiring 1.15 subtasks compared to the expert with a subtask optimality rate of 81\%.
We provide an example three-robot mission in Fig.~\ref{fig:example_real_mission}, where the planner is tasked with triaging a chemical spill. The planner correctly tasks the Husky and Spot in parallel. 
Once the chemical container is localized, the planner tasks the Jackal to deliver a chemical containment package.
Fig.~\ref{fig:method} illustrates a four-robot triaging mission. The planner tasks the UAV to build an overhead map, while the Spot performs longer range exploration, the Jackal serves as a communication relay, and the Husky traverses rugged terrain. Once relevant semantics are found (\ie a chemical barrel), the planner tasks inspection for detailed analysis.

\begin{table}[t!]
\vspace{5pt}
\centering
    \begin{tabular}{c | cccc} \toprule
    Method & Success ($\uparrow$) & ST ($\downarrow$) & Opt. ($\uparrow$) & Tokens ($\downarrow$) \\ \toprule
    Expert Planner & 100.0\% & 1.0 & 100.0\% & - \\ 
    Ours & 87.5\% & 1.15 & 81.0\% & 4.7k \\ \bottomrule
    \end{tabular}
    \caption{Real-world results across 16 missions. \vspace{-16pt}}
    \label{tab:real-results}
\end{table}

\subsection{Ablation Studies}
\label{sec:ablation}
Finally, we ablate key components of \textsc{SPINE-HT} on a subset of 6 specifications. 
We first remove the task allocation procedure described in \S\ref{sec:method_assign}, instead having the LLM directly assign robots to subtasks.
Next, the subtask generation LLM directly predicts subtasks, instead of using the generation process described in \S\ref{sec:method_decomp}.
Finally, we remove the feedback loop described in \S\ref{sec:method_feedback}, which requires the LLM to generate open-loop plans.
 
We report results in Tab.~\ref{tab:ablations}.
Without the task allocation module, subtask optimality drops from 90.2\% to 80.0\%. 
We find this is due to poor spatial reasoning by the LLM---it does not account for distance to goal.
Without reasoning for subtask generation, we observe a nearly 17\% drop in mission success with 21\% less optimal subtasks, indicating that structured generation is crucial for maximally successful and optimal planning.
Perhaps unsurprisingly, the planner cannot achieve any missions without feedback, underscoring the importance of online adaptation.

\subsection{Failure Modes}
We observed two primary failure modes.
The first was physical in nature.
In one example, the subtask generator LLM hallucinated a navigation objective, which was caught by the validator. However, the LLM could not provide a realizable plan, thus the mission terminated after reaching the maximum number of generation attempts.
The second failure mode was semantic.
For example, in one experiment a robot was not able to complete a subtask.
The LLM incorrectly interpreted subtask feedback---which indicated subtask failure---and prematurely declared the mission a success.
In actuality, the LLM should have provided a corrective subtask in order to continue the mission.
In all cases, the physical validation procedure correctly prevented unrealizable tasks. 
However, semantic validation, e.g., ensuring that the generated subtasks are aligned with the mission goals, is still an outstanding topic of research.

\begin{table}[h!]
\centering
    \begin{tabular}{c | cccc} \toprule
    Method & Success ($\uparrow$) & ST ($\downarrow$) & Opt. ($\uparrow$) & Tokens ($\downarrow$) \\ \toprule
    Ours & 100.0\% & 1.1 & 90.2\% & 3.7k \\
    - Assignment & 100.0\% & 1.2 & 80.0\% & 4.5k  \\ 
    - Reasoning & 83.3\% & 1.3 & 69.0\% & 2.5k  \\ 
    - Feedback & 0.0\% & - & -  & 14.4k  \\ \bottomrule
    \end{tabular}
    \caption{Ablation results averaged over 6 specifications.\vspace{-24pt}}
    \label{tab:ablations}
\end{table}

\section{Conclusion}
\label{sec:conclusion}
We consider the problem of heterogeneous robot collaboration given incomplete natural language specifications, where the relevant subtasks and semantics must be inferred.
We present \textsc{SPINE-HT}, a framework that addresses this problem by grounding the generative reasoning of LLMs into robot capabilities and environmental constraints via a three-stage framework that leverages online plan validation and feedback.
Through simulation and real-world experiments comprising air and ground platforms, we demonstrate \textsc{SPINE-HT}'s ability to infer grounded subtasks from natural language specifications, assign subtasks based on robot capabilities, and leverage feedback to refine subtasks online.
We envision three promising directions for future work.
First, our generation and assignment modules run centrally, and designing decentralized alternatives is a natural extension of our framework.
Developing on-device language models for robot teaming is a promising means of removing reliance on cloud-hosted LLMs and furthering the robustness of our system.
Finally, we would also like to consider settings requiring manipulation or other robot modalities.
\vspace{-3pt}

\bibliographystyle{IEEEtran}
\bibliography{IEEEabrv, refs}

\newpage

\clearpage

\appendices

% \begin{bibunit}[plain]

\setcounter{figure}{0}
\setcounter{table}{0}
\renewcommand{\thefigure}{A\arabic{figure}}
\renewcommand{\thetable}{A\arabic{table}}
\renewcommand{\thesection}{A\arabic{section}}

\section{Summary}
\label{appendix:summary}

This appendix provides additional implementation and experimental details for \texttt{SPINE-HT}. 
Section~\ref{appendix:implementation} provides implementation details on the LLM prompts (\S\ref{appendix:llm}), robot behaviors (\S\ref{appendix:behaviors}), and UAV mapper (\S\ref{appendix:uav_mapping}) used by our framework. 
Section~\ref{appendix:experimental} provides further experimental details. 
Section~\ref{appendix:missions} provides the full list of mission specifications and team configurations used in our experiments.
Section~\ref{appendix:environments} provides further details about our evaluation environments.
Sections~\ref{appendix:example_real_experiment} and \ref{appendix:example_sim_mission} walk through two experimental missions---one real and one simulated.
Finally, Section~\ref{appendix:analysis} provides further experimental analysis.

\section{Implementation Details}
\label{appendix:implementation}

\subsection{LLM Configuration}
\label{appendix:llm}

The LLM's system prompt consists of four main components: a role overview, a definition of available behaviors, the mapping API, and the planning API. 
The role overview defines the LLM's high-level objectives (Listing~\ref{listing:role}). 
The LLM is instructed to generate a plan using a predefined list of behaviors (Listing~\ref{listing:behavior}), which are described in Section~\ref{appendix:behaviors}. 
The LLM receives semantic map updates via the API defined in Listing~\ref{listing:mapping_api}. 
The final portion of the LLM prompt defines the plan generation format (Listing~\ref{listing:planning_instructions}). 
As stated in Section~\ref{sec:method_feedback}, the LLM begins the mission with input specifications from the user. 
All updates are provided in-context via the mapping API.

\subsection{Behaviors}
\label{appendix:behaviors}

We consider five ground robot behaviors and two aerial robot behaviors, which are described in Listing~\ref{listing:behavior}. 
These behaviors take grounding arguments comprising entities in the semantic map or queries, which are used to extract detailed and mission-relevant information from a VLM. 
The ground robot capability arguments (\texttt{ugv\_type}) allow the planner to specify any UGV, a platform type, or a particular robot (\texttt{robot\_name}). 
For example, if the mission requires a robot to traverse rugged terrain, the planner may invoke a navigation behavior for a Husky.
The preconditions for behaviors requiring navigation---\texttt{ugv\_navigate}, \texttt{ugv\_inspect}, \texttt{ugv\_map\_region}, and \texttt{uav\_map}---require that a valid path exists in the current semantic map.
Because we only consider one UAV in our team, the capability argument of the UAV behaviors is implied.
The \texttt{ugv\_inspect} behavior further requires that the target object is connected to a region node.
All behaviors return a boolean value that indicates whether the behavior is successful. 
Behaviors may also provide information acquired during execution, such as semantic map updates via the interface defined in Listing~\ref{listing:mapping_api}.

\subsection{UGV Mapping}
\label{appendix:ugv_mapping}
As discussed in \S V-A, the UGV semantic mapping framework takes three sources of input: odometry, a Lidar pointcloud, and RGB-D image. 
We use Direct Lidar Inertial Odometry for localization~\cite{chen2022dlio}.
Pointclouds are segmented for obstacles by GroundGrid~\cite{GroundGrid}, and these obstacles are used to determine freespace for region and edge placement.
GroundingDino performs open-vocabulary object detection~\cite{liu2023groundingdino}.
Detections are localized in 3D space, associated via a multiple-hypothesis tracker, and associated tracks are added to the semantic map~\cite{reid1979algorithm}.
Finally, we use the VIP-Llava VLM for region and object captioning~\cite{cai2023makingllava}. 

\subsection{UAV Mapping}
\label{appendix:uav_mapping}

The UAV provides a sparse map in a graphical structure as shown in Fig. \ref{fig:mapping} and Fig. \ref{fig:appendix_uav_map}. The camera module runs its own state estimate system to get accurate metric positions and orientation of the camera. We leverage GTSAM~\cite{gtsam} for this system and GPS as prior factors and preintegrate the IMU measurements. We optimize using ISAM2~\cite{isam2} every time a GPS factor is added. The IMU preintegration allows us to get state estimates at the IMU rate, 400 Hz, allowing us to get an accurate global position for each image. To do the semantic segmentation we use Grounded-SAM2~\cite{ravi2024sam2segmentimages}. The semantic classes are defined by the SPINE-HT planner and sent to the UAV. We predefine the semantic class of ``road'' as the defining class to place region nodes. When roads are detected and segmented we take the largest road segment and choose the point on the mask that is furthest from other classes, to ensure traversability by the ground robots. New region nodes are added to the closest other road node if it is sufficiently close, otherwise it is only connected to the previous region node. When objects are detected and masked we take the point on the mask that is closest to the road mask and then connect it to the closest region node. The graph is generated incrementally as more images are processed. All the nodes are georegistered since the camera's position is estimated in the ENU frame. The georegistered graph can then be sent to the ground robots.

\section{Experimental Details}
\label{appendix:experimental}

\subsection{Mission Specifications}
\label{appendix:missions}

We provide the set of missions and team configurations used in our experiments. Simulated experiments are detailed in Table~\ref{tab:appendix_sim_missions} and real-world missions are shown in Table~\ref{tab:appendix_real_missions}.
To introduce heterogeneity, some missions designate specific capabilities. 
For example, during missions that emulate chemical spill triage, one robot was specified to be carrying a chemical containment package. 
The planner must use this information to appropriately designate subtasks.

\subsection{Experimental Environments}
\label{appendix:environments}

\begin{figure}
    \centering
    \includegraphics[width=1.0\linewidth]{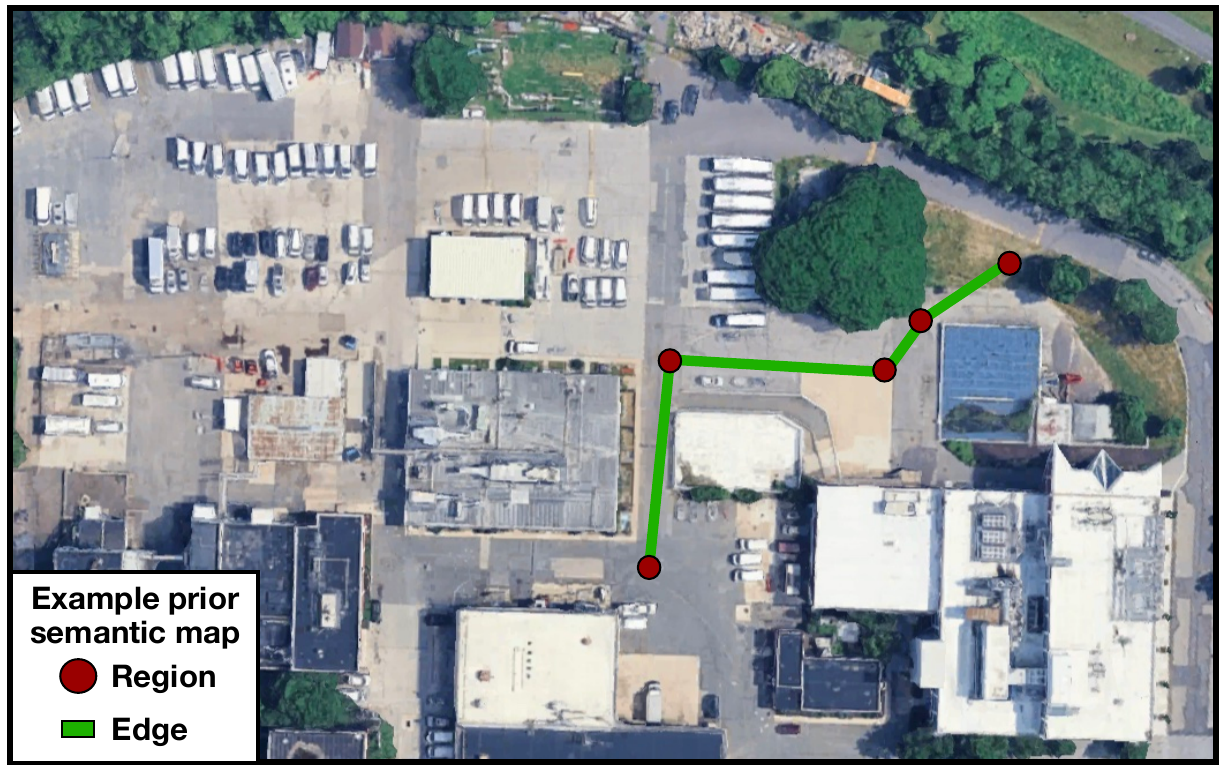}
    \caption{Top-down view of environment used for real-world experiments with an example prior map overlaid.}
    \label{fig:appendix_pennov}
\end{figure}

\begin{figure}
    \centering
    \includegraphics[width=1.0\linewidth]{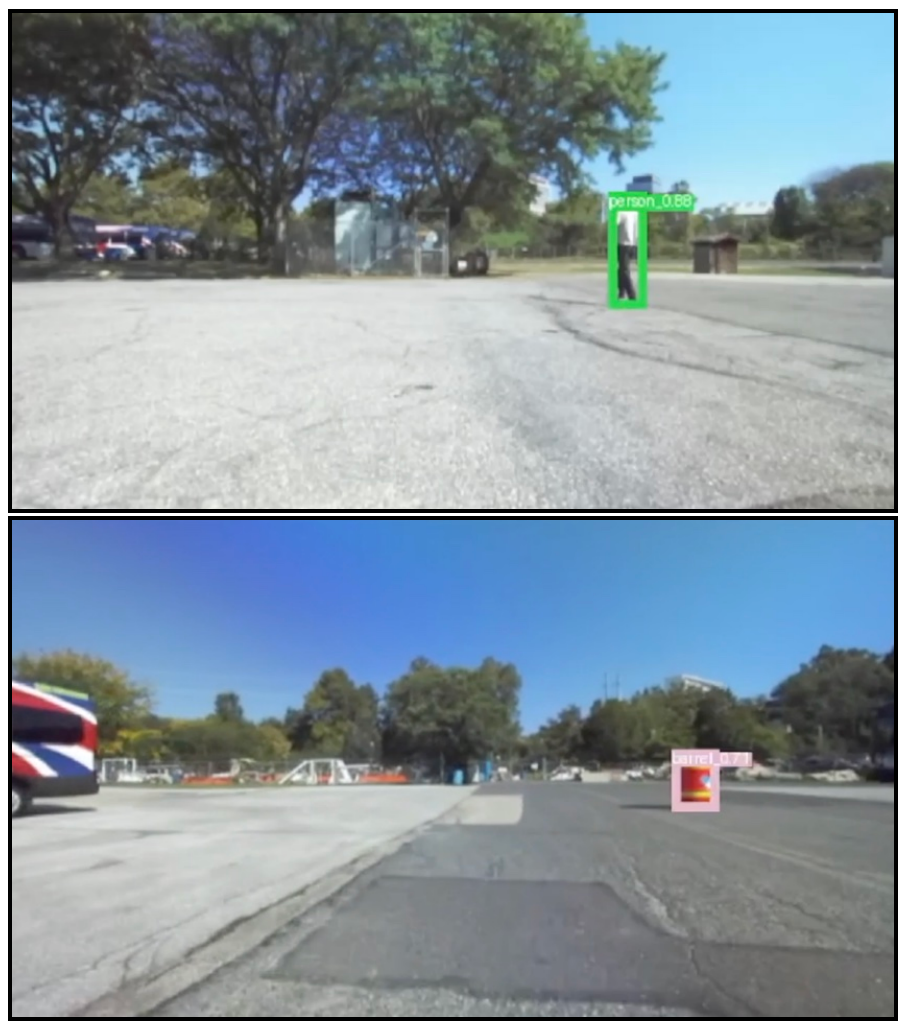}
    \caption{Semantics discovered by robots and added to their semantic map during the mission described in \S\ref{appendix:example_real_experiment}.}
    \label{fig:appendix_semantics}
\end{figure}

\begin{figure}
    \centering
    \includegraphics[width=0.95\linewidth]{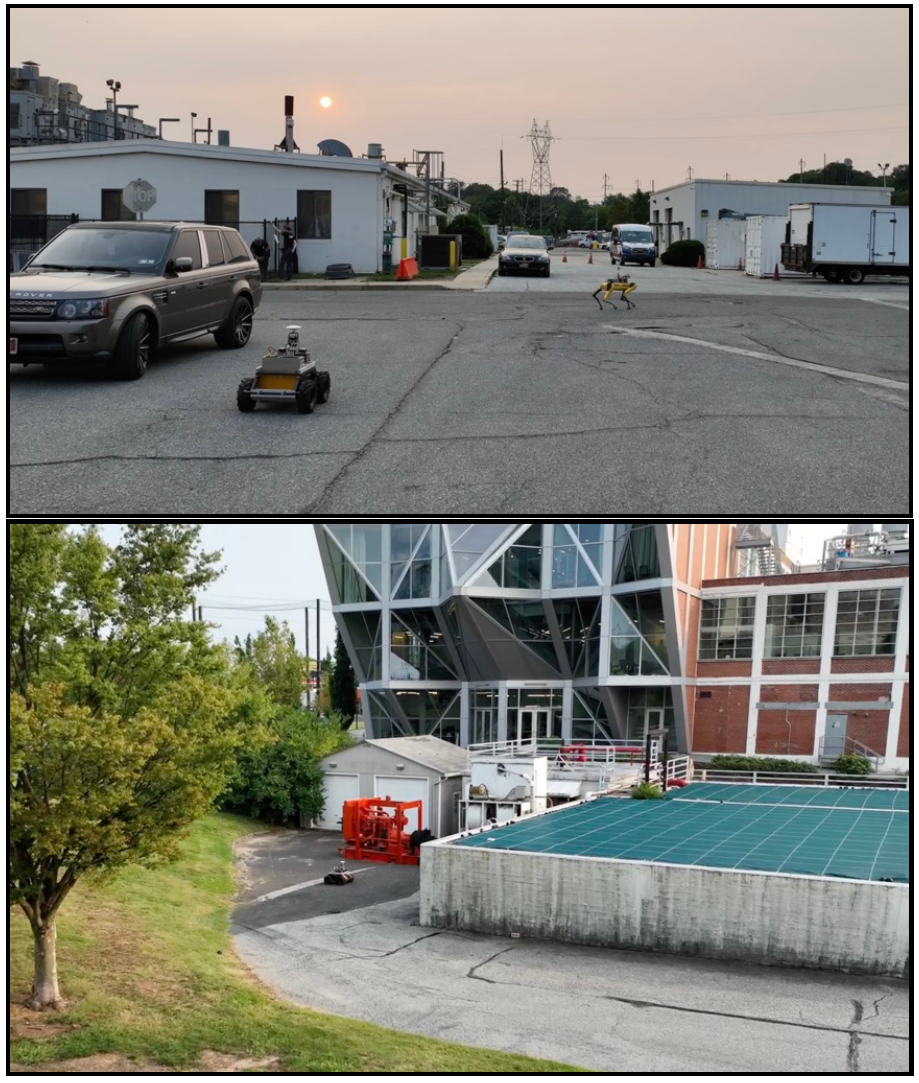}
    \caption{Views of experimental environments.}
    \label{fig:appendix_terrain}
\end{figure}

This section provides additional details about experimental environments, including semantics and prior maps used to initialize the semantic mapping framework.
We provide a top-down view of the environment used for real-world experiments in Fig.~\ref{fig:appendix_pennov}. 
The environment features a range of semantics (see Fig.~\ref{fig:appendix_semantics}) and terrain types (see Fig.~\ref{fig:appendix_terrain}), which enables complex missions.

\begin{figure}
    \centering
    \includegraphics[width=0.95\linewidth]{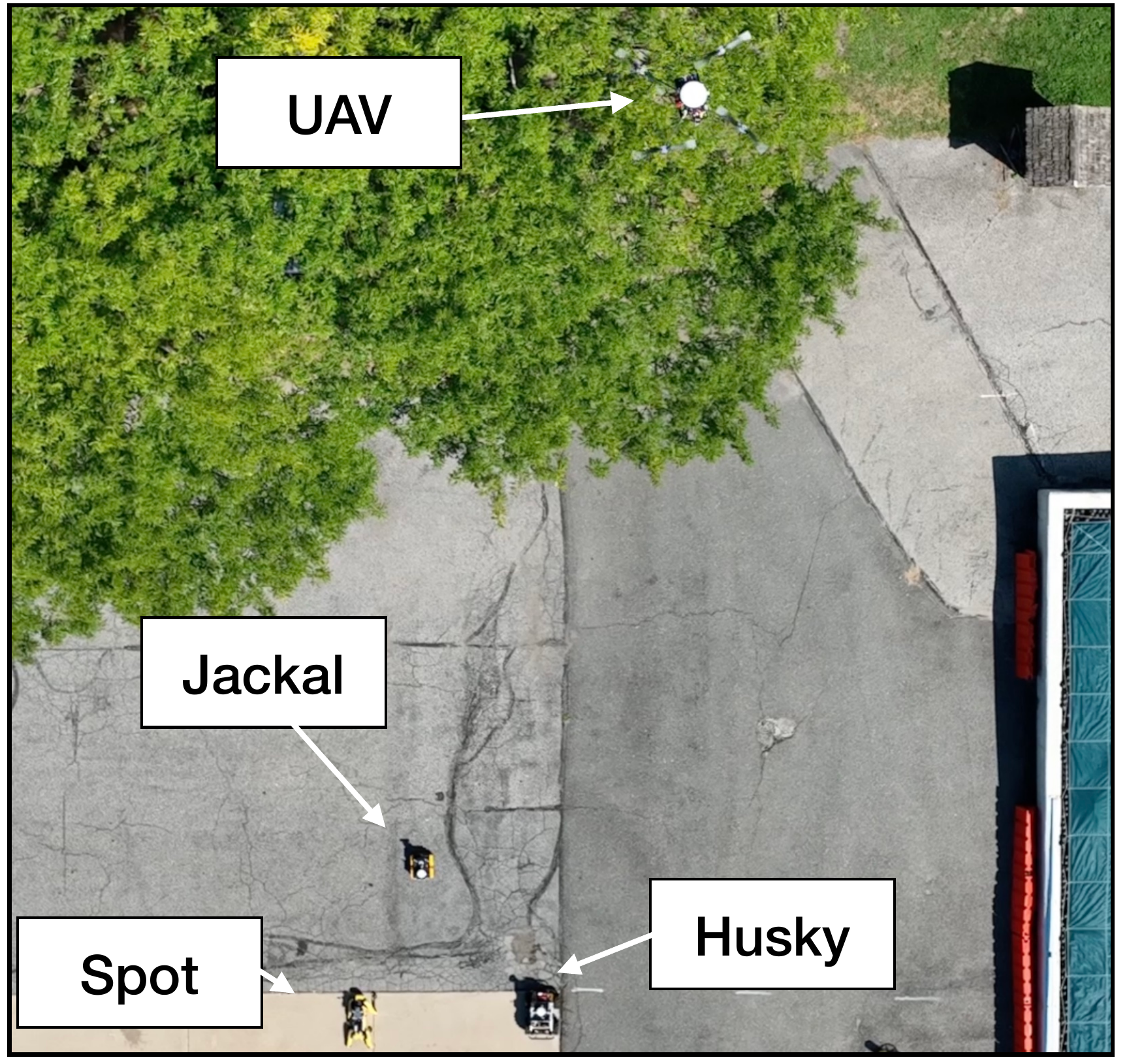}
    \caption{Robots at the beginning of the experiment described in Section~\ref{appendix:example_real_experiment}.}
    \label{fig:appendix_example_real_experiment}
\end{figure}

\medkip

\subsection{Real-World Multi-Robot Triage Mission.}
\label{appendix:example_real_experiment}
We describe a mission where \texttt{SPINE-HT} must triage a potential chemical spill using four robots: a Boston Dynamics Spot, a Clearpath Jackal, a Clearpath Husky, and a high-altitude UAV.
All robots start at the same location, as shown in Fig.~\ref{fig:appendix_example_real_experiment}.
\texttt{SPINE-HT} was given the following mission specification: \emph{``I got reports of a chemical barrel near people around (45, 100) and terrain 1. Establish comms in a central northern location, map the area, and report findings. Use the Spot to map UAV-discovered regions.''} 
The team specification was: \emph{``You have one Boston Dynamics Spot, one Clearpath Husky, one Clearpath Jackal, and one high-altitude UAV. The Jackal is equipped with a strong radio.''} 
The planner was also given the prior map illustrated in Fig.~\ref{fig:appendix_pennov} via JSON (condensed for brevity):

\begin{tcolorbox}[left=1mm, right=1.5mm, top=1.5mm, bottom=1mm] 
\footnotesize
\begin{minted}[breaklines, breakanywhere]{json}
{"regions": [
    {"name": "ground_1", "coords": [0, 0]},
    {"name": "ground_2", "coords": [0, 20]},
    {"name": "ground_3", "coords": [0, 45]},
    {"name": "terrain_1", "coords": [20, -20], 
     "description": "in rough terrain"},
    {"name": "terrain_2", "coords": [6, -35]}
],
"objects": [],
"region_connections": [
    ["ground_1", "terrain_1"],
    ["terrain_1", "terrain_2"],
    ["ground_1", "ground_2"],
    ["ground_2", "ground_3"]
],
"object_connections": []
}
\end{minted}
\end{tcolorbox}

\noindent
This provides a skeleton of the environment used to initialize planning. \texttt{SPINE-HT} tasks the Jackal to establish communication and Husky to explore the rougher terrain, while concurrently tasking the UAV to map more distant areas:

\begin{tcolorbox}[left=1mm, right=1.5mm, top=1.5mm, bottom=1mm] 
\footnotesize
\begin{minted}[breaklines, breakanywhere]{python}
uav_explore_to(x=45.0, y=100.0)
ugv_navigate(region_node='ground_3',
    ugv_type='jackal')
ugv_map_region(region_node='terrain_1', 
    ugv_type='husky')
\end{minted}
\end{tcolorbox}

\begin{figure}
    \centering
    \includegraphics[width=1.0\linewidth]{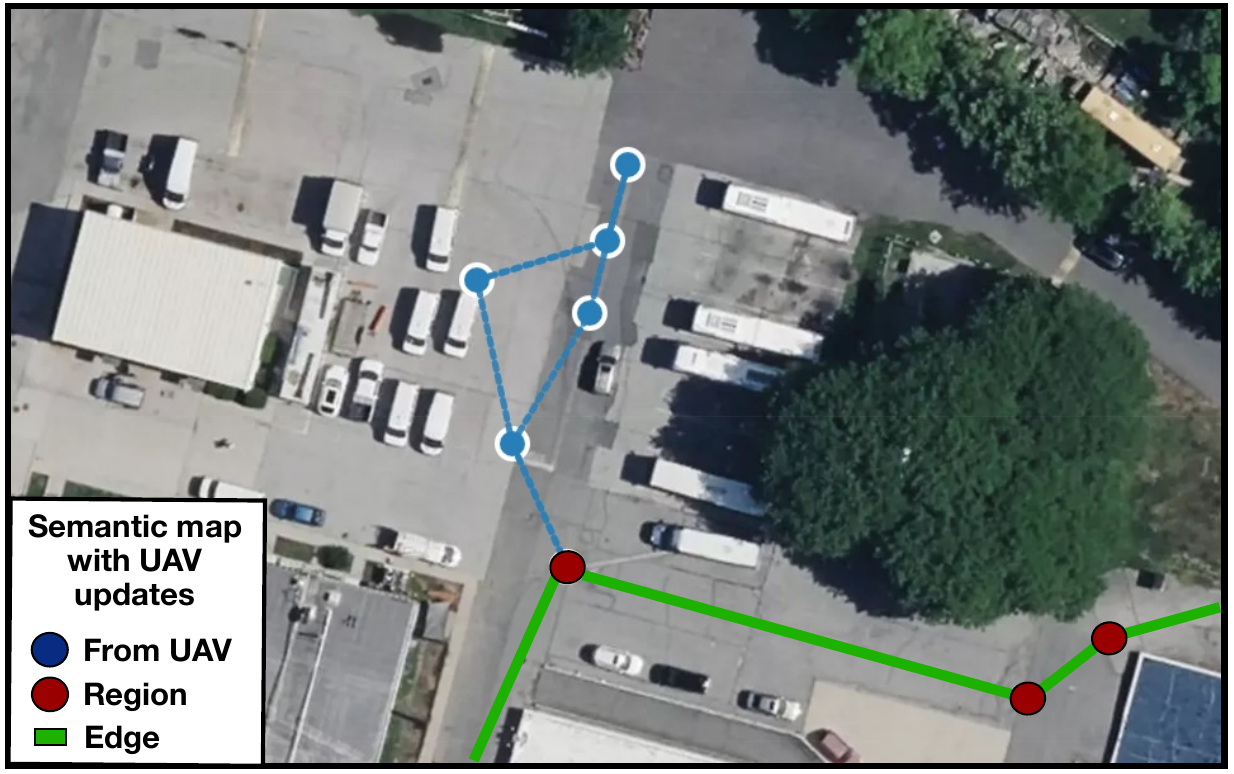}
    \caption{Semantic map comprising priors and overhead maps constructed by the UAV during the mission described in \S\ref{appendix:example_real_experiment}.}
    \label{fig:appendix_uav_map}
\end{figure}

During subtask execution, the Jackal discovered a person (illustrated in Fig.~\ref{fig:appendix_semantics}, top), the Husky successfully navigated to its goal, and the UAV built an overhead map of the mission-relevant area (Fig.~\ref{fig:appendix_uav_map}). 
These updates were provided to \texttt{SPINE-HT} via the following in-context update:

\begin{tcolorbox}[left=1mm, right=1.5mm, top=1.5mm, bottom=1mm] 
\footnotesize
\begin{minted}[breaklines, breakanywhere]{text}
jackal_1 updates: 
    add_nodes({name: discovered_person_0, 
        type: object, 
        coords: [-3.2, 24.2]}),
    add_connections(
        [discovered_person_0, ground_2]),
    update_robot_location(ground_3)

husky_1 updates: 
    update_robot_location(terrain_1)

uav_1 updates: 
    add_nodes(
        {coords: [35.4, 54.4], description: from_uav, 
        type: region, name: road_1_from_uav}, 
        {coords: [27.5, 49.9], description: from_uav, 
        type: region, name: road_2_from_uav}, 
        ...
    ),
    add_connections(
        [road_1_from_uav, road_2_from_uav], 
        [road_1_from_uav, road_2_from_uav, ...])
\end{minted}
\end{tcolorbox}

\noindent
\texttt{SPINE-HT} then tasked the Spot to explore the first UAV-discovered region and inspect the discovered person for signs of harm via the subtasks:

\begin{tcolorbox}[left=1mm, right=1.5mm, top=1.5mm, bottom=1mm] 
\footnotesize
\begin{minted}[breaklines, breakanywhere]{python}
ugv_explore_to_node(region_node=
    'road_2_from_uav', 
    ugv_type='spot')
ugv_inspect(object_node='discovered_person_0', 
    query='signs of chemical exposure or distress', 
    ugv_type='spot')
\end{minted}
\end{tcolorbox}

\noindent
During the first subtask, the Spot discovered a chemical barrel (see Fig.~\ref{fig:appendix_semantics}, bottom).
Because the mission explicitly states to discover a chemical barrel, \texttt{SPINE-HT} refined Spot's subtask sequence to prioritize barrel inspection, which Spot completed successfully and provided the mission-relevant information to the user.

\subsection{Simulated Multi-Robot Care Package Delivery Mission.}
\label{appendix:example_sim_mission}

\begin{figure}
    \centering
    \includegraphics[width=1.0\linewidth]{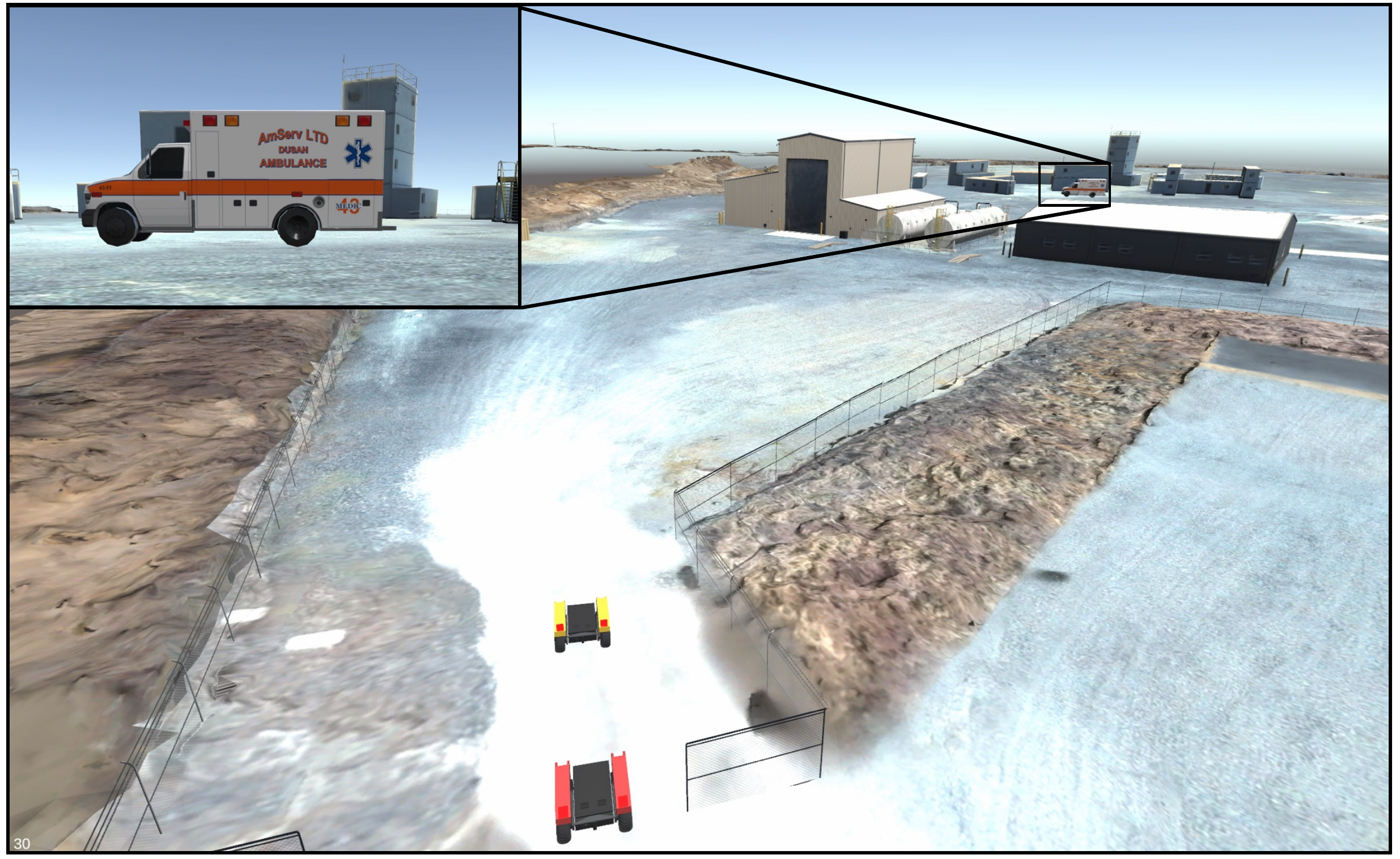}
    \caption{Simulation environment used for the experiment described in \S\ref{appendix:example_sim_mission}. A robot had to deliver a package to the ambulance, whose location was initially unknown. The other robot acted as a scout.}
    \label{fig:appendix_sim_env}
\end{figure}

This mission requires \texttt{SPINE-HT} to deliver a care package to an ambulance whose location is unknown.
\texttt{SPINE-HT} has access to two Clearpath Warthog robots---\texttt{Warty} and \texttt{Wanda}.
Our simulator only provides Clearpath Warthog models, so we introduce heterogeneity by providing individual robots with unique traits. 
In this mission, 
\texttt{Warty} is carrying a heavy care package that needs to be delivered to the ambulance once it's found. 
\texttt{SPINE-HT} was given the mission specification: \emph{``An ambulance has been reported at one of the building sites. Find it and deliver the package.''}, and the team specification \emph{``You have two clearpath warthog robots named warty and wanda. warty is carrying a heavy care package and hence moves slower than wanda.''} 
The planner was given a prior map of the environment shown in Fig.~\ref{fig:appendix_sim_env} as follows (condensed for brevity):

\begin{tcolorbox}[left=1mm, right=1.5mm, top=1.5mm, bottom=1mm] 
\footnotesize
\begin{minted}[breaklines, breakanywhere]{json}
{"regions": [
    {"name": "region_1", "coords": [0, 0]},
    {"name": "region_2", "coords": [40, -20]},
    {"name": "region_3", "coords": [70, -27]},
    {"name": "region_4", "coords": [53, -101]},
    {"name": "region_5", "coords": [113, -118]}
],
"objects": [
    {"name": "building_1", "coords": [91, -40]},
    {"name": "building_2", "coords": [134, -127]}
],
"region_connections": [
    ["region_1", "region_2"],
    ["region_2", "region_3"],
    ["region_2", "region_4"],
    ["region_4", "region_5"]
],
"object_connections": [
["region_3", "building_1"],
["region_5", "building_2"]
]}
\end{minted}
\end{tcolorbox}

\noindent \texttt{SPINE-HT} tasks the slower robot, \texttt{Warty}, to inspect the area around the closer building (\texttt{building\_1}) and the faster robot, \texttt{Wanda}, to inspect the farther building (\texttt{building\_2}):
\begin{tcolorbox}[left=1mm, right=1.5mm, top=1.5mm, bottom=1mm] 
\footnotesize
\begin{minted}[breaklines, breakanywhere]{python}
ugv_map_region(region_node='region_3', ugv_type='warty')
ugv_map_region(region_node='region_5', ugv_type='wanda')
\end{minted}
\end{tcolorbox}

\noindent The ambulance is parked near \texttt{building\_2}.
\texttt{Wanda} finds it, adds it to the semantic map, and notifies \texttt{SPINE-HT} via an in-context update:

\begin{tcolorbox}[left=1mm, right=1.5mm, top=1.5mm, bottom=1mm] 
\footnotesize
\begin{minted}[breaklines, breakanywhere]{text}
Wanda updates: 
add_nodes({name: discovered_ambulance_0, type: object, 
         coords: [123, -121]}),
add_connections([discovered_ambulance_0, region_5]),
update_robot_location(region_5)

Warty updates: 
  update_robot_location(region_3)
\end{minted}
\end{tcolorbox}

\noindent Having discovered the ambulance, \texttt{SPINE-HT} tasks \texttt{Warty}---the robot carrying the package---to the ambulance. 

\section{Further Experimental Analysis}
\label{appendix:analysis}

This section provides further experimental analysis and discusses promising future directions.

\subsection{Comparative Performance.}

As described in Section \ref{sec:sim_result}, we run the baseline planner~\cite{liu2025coherentcollaborationheterogeneousmultirobot} on the same set of simulated missions as \texttt{SPINE-HT}, and we observe three main limitations with the baseline planner. 
Firstly, the baseline planner generates sequential instead of parallelized tasking of robots. 
For example, at each planning step, the baseline planner will assign only one robot to traverse a single region node in the semantic map, even though it has access to the entire graph and can task robots in parallel directly to the goal location. 
In contrast, \texttt{SPINE-HT} parallelizes tasking, which leads to more efficient mission execution as shown in Tab.~\ref{tab:sim-results}. Secondly, the baseline planner struggles to understand the appropriate termination conditions when given a mission with incomplete specifications.
For example, the \emph{establish communication} missions listed in Tab.~\ref{tab:appendix_sim_missions} require a robot to establish communication by transporting a radio to a particular location. 
Although the baseline planner provides an initially coherent plan, it does not recognize when it has met the termination conditions of guiding a radio-equipped robot to the goal location. 
\texttt{SPINE-HT}, on the other hand, understands that sending the robot to a goal location establishes communications.
Lastly, the baseline planner often doesn't understand the mission-relevant implications of certain team specifications. 
For example, in the care package mission described in \S\ref{appendix:example_sim_mission}, the baseline planner infers that having a heavy care package means the robot should not move at all until the ambulance is found. 
The optimal task allocation, however, is to assign the heavier, slower-moving robot to the nearer location and the lighter, faster-moving robot to the farther location, which \texttt{SPINE-HT} accomplishes.

\subsection{Towards a Tool Use Paradigm.}

The ability to translate high-level semantic goals into grounded robot plans is a core strength of our framework, and this ability is enabled by LLM reasoning. 
We observe, however, LLMs may struggle to reason about lower-level information such as spatially-optimal task assignments, as demonstrated in our ablation studies~\S V-D. For example, in one of the simulated missions, the robots were tasked with communicating with each other, but they had to be \emph{within a 30-meter radius} of each other to do so. In situations like these, the planner ignores this spatial constraint in the mission specification and still facilitates communication when robots are beyond range.
We envision that effective instantiations of our framework will adhere to a ``tool use'' paradigm wherein an LLM calls high-level behaviors.
For example, several of our experimental missions require a robot to act as a communication relay. 
Instead of having \texttt{SPINE-HT} directly place robots, future work may leverage existing functionality for efficient communication~\cite{cladera2024enabling, gielis2022critical, gil2015adaptive} or exploration~\cite{ong2025atlasnavigatoractivetaskdriven}.

\newpage

\begin{listing*}[h!]
\caption{Role overview}
\label{listing:role}
\begin{tcolorbox}[left=1mm, right=1.5mm, top=1.5mm, bottom=1mm] 
\footnotesize
\begin{minted}[breaklines, breakanywhere]{text}
# Role and Objective
- Serve as a multi-robot task allocator, generating and refining task allocation plans for a fleet of robots. Respond dynamically to changes in team composition, mission objectives, and the semantic graph environment.

# Instructions
- Begin each planning iteration with a concise conceptual checklist (3-7 bullets) 
    - Outlining key planning steps (avoid implementation specifics).
    - Identifying critical enabling tasks (e.g., network setup, access clearing) and ensure they precede dependent tasks.
- For each step, input includes: <team specification>, <mission specification>, and <semantic graph>.
- Explicitly identify and list all mission-relevant regions and objects by aligning mission terms to nodes in the semantic graph. 
    - Always attempt to ground mission concepts to existing nodes before generating tasks. Start with the most relevant nodes.
    - If multiple candidate nodes exist, explain the mapping choice.
    - If no relevant nodes exist, state this clearly and fall back to exploration or prerequisite mapping tasks.
- Produce a <team allocation> as a single JSON object, conforming strictly to the output schema below. Ensure all field specifications and planning constraints are followed exactly.
- Planning is iterative: after each plan execution, you may receive updated feedback or semantic graph modifications. Revise and regenerate your plan in each new iteration.

## Robot team specification
- {team_specification}
- Available robot APIs are detailed below; only specify robot type in task calls if required by mission requirements.
- Pay attention to environment semantics and robot capabilities when tasking heterogeneous robots. Some robots may be better at traversing difficult terrain, have more payload capacity, etc.
- Here are the general robot capabilities
    - Husky: Can traverse rugged terrain and is good for finding new paths, however it is slow.
    - Spot: Is fast and good for exploration
    - Jackal: A middleground between the husky and spot

## Robot APIs
- Each robot function is defined below; function signatures and docstrings specify expected parameters and return fields.
- Assign a specific robot only when the task requires its unique capability. Otherwise, use "any" to indicate any available robot can perform the task.
When a task explicitly requires a unique capability (e.g., best radio), specify the exact robot ID (e.g., Jackal_2).
- Try to task all robots, unless dependency conditions are stated in the mission
- For APIs with a robot_name option, specify a particular robot only if there is a good reason (i.e., a capability requirement).
\end{minted}
\end{tcolorbox}
\end{listing*}

\newpage

\begin{listing*}[h!]
\caption{Behavior description}
\label{listing:behavior}
\begin{tcolorbox}[left=1mm, right=1.5mm, top=1.5mm, bottom=1mm] 
\footnotesize
\begin{minted}[breaklines, breakanywhere]{python}
def ugv_navigate(
    region_node: str, ugv_type: str = ["any", "jackal", "husky", "robot_name"]
) -> bool:
    """Navigate to a node in the semantic graph. This MUST refer to an existing node.

    Parameters
    ---
    region_node: str
        Existing node in the graph
    ugv_type: str
        Pick most appropriate option (jackal, husky, or any)
    """

def ugv_inspect(
    object_node: str, query: str, ugv_type=["any", "jackal", "spot", "robot_name"]
) -> str:
    """Navigate to the closest region, then inspect `object_node` for a specific attribute `query`.
    The query is processed by a vision language model (VLM), and the VLM's answer will be provided.
    This subsumes navigating to the region closest to the object.
    """

def ugv_map_region(
    region_node: str, ugv_type: str = ["any", "jackal", "husky", "spot", "robot_name"]
) -> str:
    """Navigate to `region_node,` gather a semantic description, and discover nearby objects.
    This subsumes navigation to region_node"""

def ugv_explore_to_coord(
    x: float, y: float, ugv_type: str = ["any", "jackal", "husky", "spot", "robot_name"]
) -> str:
    """Try to navigate to the coordinate (x, y). 
    If successful, you will update your semantic map with a node at this location.
    Note that you may not reach the coordinate exactly. Allow for about 10 meters of error.
    
    Use this if the mission requires navigating to a coordinate that is not close (within 10 meters) to an existing node.
    """

def ugv_explore_to_node(region_node: str, ugv_type=["any", "jackal", "husky", "spot", "robot_name"]) -> str:
    """Attempt to find a path to `region_node` if none exists.

    ONLY call this if there is no existing path"""

def uav_map(region_node: str) -> bool:
    """Fly to a node in the graph and map that area. Returns map updates"""

def uav_explore_to(x: float, y: float) -> str:
    """Explore an (x, y) coordinate to expand the semantic map. Returns map updates."""
\end{minted}
\end{tcolorbox}
\end{listing*}

\newpage

\begin{listing*}[h!]
\caption{Mapping API}
\label{listing:mapping_api}
\begin{tcolorbox}[left=1mm, right=1.5mm, top=1.5mm, bottom=1mm] 
\footnotesize
\begin{minted}[breaklines, breakanywhere]{python}
## Semantic Graph
- Provided as a JSON object with fields: objects, regions, object_connections, region_connections, and init_location. Example:

{{
"objects": [{{"name": "object_1_name", "coords": [x, y]}}],
"regions": [{{"name": "region_1_name", "coords": [x, y]}}],
"object_connections": [["object_name", "region_name"]],
"region_connections": [["region_name_a", "region_name_b"]],
"init_location": "starting_region"
}}

- Graph updates are given via API:

def remove_nodes(removed_nodes: List[str]) -> None:
    """Remove `nodes` and associated edges from graph."""


def add_nodes(new_nodes: List[Dict[str, str]]) -> None:
    """Add nodes to graph. Each node is represented as a dictionary."""


def add_connections(new_connections: List[Tuple[str, str]]) -> None:
    """Add a list of connections. Each element is a tuple of the connecting nodes."""


def remove_connections(removed_connections: List[Tuple[str, str]]) -> None:
    """Removes a list of connections. Each element is a tuple of the endpoint nodes in the edge."""


def update_robot_location(region_node: str) -> None:
    """Update robot's location in the graph to `region_node`."""


def update_node_attributes(attribute: List[Dict[str, str]]) -> None:
    """Update node's attributes. Each entry of the input is a dictionary of new node values.
    Entries will include the referent node's name."""


def no_updates() -> None:
    """There have been no updates."""
\end{minted}
\end{tcolorbox}
\end{listing*}

\newpage

\begin{listing*}[h!]
\caption{Planning instructions}
\label{listing:planning_instructions}
\begin{tcolorbox}[left=1mm, right=1.5mm, top=1.5mm, bottom=1mm] 
\footnotesize
\begin{minted}[breaklines, breakanywhere]{text}
# Planning and Verification
- In each mission cycle:
- Ensure that all proposed tasks are executable within the current semantic graph and robot team.
- Before returning output, verify all tasks for correctness and completeness.

# Make concise plans
- Do not call subsumed tasks. For example, if inspecting an object subsumes navigation, so do not call navigation then inspection unless necessary

# Feasibility and Adapting plans
- Always address infeasibility or feedback by updating your plan, such as generating intermediate subtasks or correcting syntax.
- Do not call tasks until they are feasible. If you receive feedback about a task, it is infeasible. Remove that from your plan and add it during a later iteration.
- Revise in response to feedback or infeasibility by correcting errors or adding intermediate plans; regenerate output.

# Extending plans
- Treat the previous plan as cumulative and authoritative unless instructed to modify otherwise; do not omit or deduplicate executed/planned tasks unless explicitly told to do so.
- Each output iteration should be a superset of the previous plan unless instructed to remove tasks.
- You will be given a list of previously completed tasks. Do not duplicate tasks during successive planning iterations. If the task did not return the designed information, calling that task again will not help.

# Sub-categories
- Use the UAV to explore and map far away locations. Do not use the UAV to map somewhere a UGV can map.
- Ignore malformed inputs; assume all provided data is in the required format.

# Verbosity
- Output a clear, concise JSON matching the specified schema. Include succinct justifications for planning decisions.
- In `tasks`, use explicit parameter names and formatting per the function signatures.
- Preserve field order and types as per the provided specification.

# Stop Conditions
- Terminate planning only upon producing a non-blank `mission_answer` indicating completion.
- In all other cases, extend the plan as a superset of the prior iteration, unless told otherwise.
- Stop once you complete all reasonable tasks and report your findings (even if there are none). Do not repeat the same plan multiple times in a row.

## Output Format
Each response must be a single JSON object with these required fields (names, types, order required):
```json
{{
"conceptual_checklist": ["steps as outlined above"],
"robot_team": "summarize robots and their abilities",
"reasoning": "string: concise rationale for the planning step",
"relevant_regions": ["most_relevant", "second_most_relevant", ...],  // This may include regions you plan to leverage in future planning iterations
"grounding_explanation": "string: explain how mission terms were mapped to semantic graph nodes, or why fallback was required",
"tasking_explanation": "Justify why you tasked robots, given their capabilities",
"relevant_graph": "List relevant portions of the graph, and explain why they are important",
"mission_answer": "string: completed mission output or blank if still planning",
"tasks": ["string", ...],
"dependency_reasoning": "string: explanation of task dependencies",
"dependency_graph": [["prior_task", "dependent_task"]],
"is_extended": true if mission is extended
}}
```

- Include all fields in every output.
- The `tasks` array must consist solely of formatted robot API call strings per the given function signatures.
- The `dependency_graph` array must contain ["prior_task", "dependent_task"] pairs exactly matching `tasks` entries. Dependencies must describe mission ordering ONLY for feasible tasks. If a task is infeasible, call it in a later iteration when it becomes feasible.
- If validation errors are found post-output, revise and redo the full JSON object, adhering strictly to this schema.

# Contextual reasoning
- Attend to semantic relationships between the mission specification and graph.
    - Make educated inferences. For example, if asked to find a car, look near the roads. Boats are near docks, etc.
\end{minted}
\end{tcolorbox}
\end{listing*}

\newpage
\clearpage

\renewcommand{\arraystretch}{1.3}
\rowcolors{2}{gray!10}{white}

\begin{table*}[h!]
\begin{tabularx}{\textwidth}{>{\raggedright\arraybackslash}X >{\raggedright\arraybackslash}X}
\toprule
\textbf{Mission specifications}  & \textbf{Platforms and capability description} \\
\midrule
There are reports of a traffic blockade under the overpass. Report what is going on. & Warthog (x2) \\
There are reports of an emergency situation at the office building. Report what is going on. & Warthog (x2) \\
There is a fire reported at the back of the factory building. Report the situation and block the front of the factory for safety. & Warthog (x2) \\
There is a fire reported at the back of the factory building. Report the situation and block the front of the factory for safety. & Warthog (x2). Only one of them is equipped with a capable front-end system for object detection and scene understanding. \\
Communications are down between the factory and HQ building. Reestablish comms and report the situation. & Warthog (x2). Both are equipped with radio relays used to establish communication. \\
Important cargo needs to be delivered from the container to the factory. Ensure there are no people or vehicles in the way. & Warthog (x2) \\
There are reports that a tsunami has struck the area. Survey the docks and report any damages to the ships. & Warthog (x2) \\
Check the front gate for signs of an intruder. Also, report the situation at the office. & Warthog (x2). One Warthog has a speaker system to sound an alarm when needed. \\
An ambulance has been reported at one of the building sites. Find it and deliver the package. & Warthog (x2). One Warthog carries a heavy care package and hence moves slower than the other one. \\
There are reports that a tsunami has struck the area. Survey the docks and report any damages to the ships. & Warthog (x3) \\
There is a chemical spill reported at the factory. Investigate the spill, go report the situation to HQ and block the front gate for safety. & Warthog (x3) \\
There is a chemical spill reported at the factory. Investigate the spill, go report the situation to the office and block the front gate for safety. & Warthog (x3). One Warthog is equipped with a containment device to capture any harmful chemicals. \\
There is a chemical spill reported at the factory. Investigate the spill, go report the situation to the office and block the front gate for safety. & Warthog (x3). Warthog 1 is equipped with a containment device to capture any harmful chemicals. Warthog 3 is equipped with a communication radio capable of communicating within a 30 meter radius. \\
Communications are down between the office, HQ and the factory. Help reestablish comms. & Warthog (x3). All three are equipped with radio relays and can communicate with each other. Warthog 3's radio device is of a much higher capacity and can communicate farther than Warthog 1's or Warthog 2's. \\
An intruder has entered from the front gate and was last sighted at the factory. See if that’s true, block the front gate for safety, and report the situation at the office. & Warthog (x3). Warthog 1 and Warthog 3 have a speaker system to sound an alarm when needed. \\
There are reports that a tsunami has struck the area. Survey the docks and report any damages to the ships. & Warthog (x4) \\
\bottomrule
\end{tabularx}
\caption{Simulated mission specifications and platform descriptions. Several specifications were repeated, yielding 24 missions.}
\label{tab:appendix_sim_missions}
\end{table*}

\newpage
\clearpage

\begin{table*}[h!]
\begin{tabularx}{\textwidth}{>{\raggedright\arraybackslash}X >{\raggedright\arraybackslash}X}
\toprule
\textbf{Specifications}  & \textbf{Platforms and capability description} \\
\midrule
Inspect two regions. & Jackal (x2) \\
Obtain the package and watch for oncoming traffic. & Jackal (x2). One jackal has a larger payload capacity. \\
Map the northmost part of the scene. & Jackal (x2) \\
Pickup the package and watch for oncoming cars. & Jackal, Spot. The Spot has a larger payload capacity. \\
Mapping. & Jackal (x2), Spot  \\
Inspect key infrastructure. & Husky, Spot. The Spot is faster while the Husky can traverse more rugged terrain. \\
Inspect key infrastructure and establish communications. & Husky, Spot, Jackal. The Spot is the fastest robot in the team, the Husky can traverse rugged terrain, while the Jackal is equipped with a powerful radio.\\
Map the northmost scene. & Husky, Spot, Jackal. The Spot is the fastest robot in the team, the Husky can traverse rugged terrain, while the Jackal is equipped with a powerful radio.\\
Inspect the chemical barrel and deliver containment package. & Husky, Spot. The Spot is the fastest robot on the team, while the Husky carries a containment package. \\
Find the chemical barrel and deliver the containment package.  & Husky, Spot. The Spot is the fastest robot on the team, while the Husky carries a containment package. \\
Inspect the chemical barrel, deliver the containment package, and block road to oncoming traffic. & Husky, Spot, Jackal. The Spot is the fastest robot on the team, the Husky can traverse rugged terrain, while the Jackal carries a containment package.\\
Find the chemical barrel, deliver the containment package, and prevent oncoming traffic. & Husky, Spot, Jackal. The Spot is the fastest robot on the team, the Husky can traverse rugged terrain, while the Jackal carries a containment package. \\
Find the chemical barrel, deliver the containment package, and block road to oncoming traffic. & Husky, Spot, Jackal. The Spot is the fastest robot on the team, the Husky can traverse rugged terrain, while the Jackal carries a containment package. \\
Establish communications, find the chemical barrel, then deliver the containment package. & Husky, Spot, Jackal. The Spot is the fastest robot on the team, the Husky can traverse rugged terrain, while the Jackal carries a containment package.\\
I got reports of a chemical spill. Triage the area.  & Husky, Spot, Jackal, UAV. The Spot is the fastest robot on the team, the Husky can traverse rugged terrain, the Jackal has moderate speed and traversability capabilities but is equipped with a semantic mapping framework, while the UAV can construct semantic maps from overhead imagery. \\
I got reports of a missing person. Triage the area.  & Husky, Spot, Jackal, UAV. The Spot is the fastest robot on the team, the Husky can traverse rugged terrain, the Jackal has moderate speed and traversability capabilities but is equipped with a semantic mapping framework, while the UAV can construct semantic maps from overhead imagery.\\
\bottomrule
\end{tabularx}
\caption{\centering Real robot missions and platform descriptions.}
\label{tab:appendix_real_missions}
\end{table*}

\newpage
\clearpage

% \putbib[refs]
% \end{bibunit}

\end{document}